\documentclass{article} 
\usepackage{iclr2025_conference,times}

\usepackage{amsmath,amsfonts,bm}

\def\eqref#1{equation~\ref{#1}}

\def\1{\bm{1}}

\DeclareMathAlphabet{\mathsfit}{\encodingdefault}{\sfdefault}{m}{sl}
\SetMathAlphabet{\mathsfit}{bold}{\encodingdefault}{\sfdefault}{bx}{n}

\usepackage{hyperref}
\usepackage{graphicx}      
\usepackage{wrapfig}
\usepackage{cleveref}
\usepackage{microtype}
\usepackage{booktabs}
\usepackage{floatrow}
\usepackage{url}
\hypersetup{
    colorlinks,
    linkcolor={red!50!black},
    citecolor={blue!50!black},
    urlcolor={blue!80!black}
}
\usepackage{algorithm}
\usepackage{algpseudocode}
\usepackage{multirow} 
\usepackage{booktabs} 

\iclrfinalcopy
\newif\ifdraft
\draftfalse

\ifdraft
    \newcommand{\yossi}[1]{{\color{orange}\textbf{Yossi:} #1}}
    \newcommand{\anish}[1]{{\color{blue}\textbf{Anish:} #1}}
    \newcommand{\nick}[1]{{\color{green!75!black}\textbf{Nick:} #1}}
    \newcommand{\ky}[1]{{\color{purple}\textbf{Kayo:} #1}}
\else
    \newcommand{\yossi}[1]{}
    \newcommand{\anish}[1]{}
    \newcommand{\nick}[1]{}
    \newcommand{\ky}[1]{}
\fi

\title{Interpreting and Editing Vision-Language \\ Representations to Mitigate Hallucinations}

\author{Nick Jiang\thanks{Equal contribution as first authors.}\hspace{0.5em}, Anish Kachinthaya\footnotemark[1]\hspace{0.5em}, Suzie Petyrk\thanks{Equal contribution as last authors.}\hspace{0.5em},Yossi Gandelsman\footnotemark[2] \\
University of California, Berkeley\\
\texttt{\{nickj,anishk,spetryk,yossi\_gandelsman\}@berkeley.edu}
}

\begin{document}

\maketitle

\begin{abstract}
We investigate the internal representations of vision-language models (VLMs) to address hallucinations, a persistent challenge despite advances in model size and training. We project VLMs' internal image representations to their language vocabulary and observe more confident output probabilities on real objects than hallucinated objects. We additionally use these output probabilities to spatially localize real objects. Building on this approach, we introduce a knowledge erasure algorithm that removes hallucinations by linearly orthogonalizing image features with respect to hallucinated object features. We show that targeted edits to a model's latent representations can reduce hallucinations by up to 25.7\% on the COCO2014 dataset while preserving performance. Our findings demonstrate how a deeper understanding of VLMs' latent representations can enhance reliability and enable novel capabilities, such as zero-shot segmentation.\footnote{Code: \url{https://github.com/nickjiang2378/vl-interp}}
\end{abstract}


\section{Introduction}


Vision-Language Models (VLMs) have recently emerged as powerful tools for understanding images via text~\citep{dai2023instructblipgeneralpurposevisionlanguagemodels,liu2024improvedbaselinesvisualinstruction}. They have demonstrated remarkable capabilities across multimodal tasks such as image captioning~\citep{li2023blip2bootstrappinglanguageimagepretraining}, visual question answering~\citep{ye2023mplugowl2revolutionizingmultimodallarge}, 
and complex multimodal reasoning~\citep{bai2023qwenvlversatilevisionlanguagemodel}. Despite their capabilities, VLMs tend to hallucinate content that does not appear in the images~\citep{Ji_2023}, which poses serious concerns for the reliability of these models in real-world applications~\citep{hu2023rsgptremotesensingvision,luo2024codisbenchmarkingcontextdependentvisual}.


Widespread belief has been that scaling to larger models and more training data will naturally mitigate hallucinations. However, recent studies have shown that hallucinations persist even in larger and more advanced models \citep{rohrbach2019objecthallucinationimagecaptioning,li2023evaluatingobjecthallucinationlarge}, suggesting that this issue cannot be solved by scale alone. Current methods reduce hallucinations by applying external interventions (e.g. object detectors; \cite{yin2023woodpeckerhallucinationcorrectionmultimodal}) or additional model fine-tuning (e.g. on hallucination examples; \cite{zhou2024analyzingmitigatingobjecthallucination,zhang2024reflectiveinstructiontuningmitigating}). 
Nevertheless, these methods often struggle to distinguish between subtle hallucinations and existing details, requiring new models or updated model parameters. 


In this paper, we aim to introduce fine-grained edits directly to the image latent representations of VLMs to reduce hallucinations without hindering their performance, an approach that has had some success in large language models ~\citep{zhang2024truthx, rutte2024latentspace}. To edit the latent representations of VLMs, we first explain their role via text. We employ the logit lens technique~\citep{nostalgebraist2020logitlens} to directly interpret the spatial VLM \textit{image} representations with VLM \textit{text vocabulary}. Surprisingly, the characteristics of these image representations are different for real objects that appear in the image and objects that are hallucinated. Moreover, the logit lens enables spatially localizing objects within the input image.




Relying on the ability to detect hallucinated objects, we edit them out by intervening in their internal representations. We introduce a knowledge erasure algorithm, \textsc{ProjectAway}, to target and remove objects by linearly orthogonalizing image features with respect to the text features of target objects. We find that \textsc{ProjectAway} can erase both real and hallucinated objects with high rates of removal. 


\begin{figure}[t!]
    \centering
    \includegraphics[width=\textwidth]{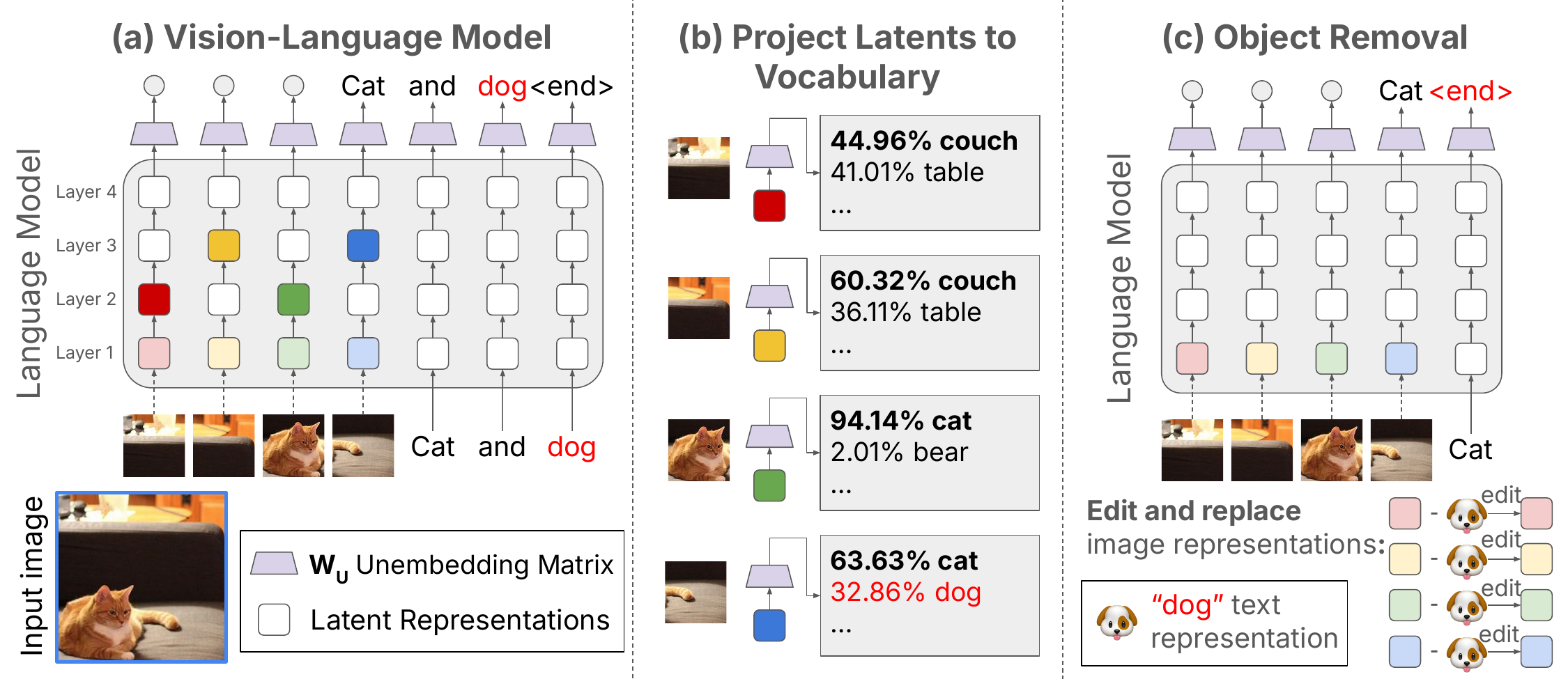}
    \vspace{-1em}
    \caption{\textbf{Interpreting VLM internal image representations}. (a) Given a VLM, (b) we unembed the latent representations from image embeddings to the vocabulary and classify hallucinations. We remove hallucinations by (c) linearly editing them out of the latent representations.}
    \label{figure:logit_and_softmax}
\end{figure}

We use our interpretation and editing approach for three tasks. First, we utilize the logit lens on image features to detect hallucinations in the image. We find that it improves mAP by 22.45\% and 47.17\% in two VLMs. Then, we combine our editing and detection method to erase hallucinations from the VLM's internal representations, reducing hallucinations up to 25.7\% on standard benchmarks, while preserving accuracy in image captioning. Finally, we use the logit lens to localize objects in the image features. We find that our spatial mapping provides comparable performance to state-of-the-art zero-shot segmentation methods. Our results indicate that understanding the internal representations of VLMs can be achieved and used to repair model hallucinations and introduce new capabilities. 

\section{Related work}

\subsection{Interpreting Latent Representations in Language Models}
\label{sec:language_interpretability}

Interpreting the inner workings of large language models enables fine-grained improvement of the language model behavior. Recent work involves utilizing the model's attention maps~\citep{kobayashi-etal-2020-attention,chefer2021genericattentionmodelexplainabilityinterpreting}, activation patterns~\citep{conmy2023automatedcircuitdiscoverymechanistic,meng2023locatingeditingfactualassociations,bronzini2024unveilingllmsevolutionlatent}, and latent representations~\citep{ghandeharioun2024patchscopesunifyingframeworkinspecting,cunningham2023sparseautoencodershighlyinterpretable,bricken2023monosemanticity} to understand their behavior with applications such as early exiting~\citep{halawi2024overthinkingtruthunderstandinglanguage} and editing or erasing the model's knowledge~\citep{dai2022knowledgeneuronspretrainedtransformers,ravfogel2024linearadversarialconcepterasure}. One class of methods probe the VLMs knowledge with linear classifiers~\citep{hewitt-manning-2019-structural,tucker2021modifiedthatsyntacticinterventions,li2024emergentworldrepresentationsexploring,belrose2023elicitinglatentpredictionstransformers}. The logit lens method~\citep{nostalgebraist2020logitlens}, which we will use in our analysis, finds the output distribution over the vocabulary of the language model at intermediate layers with the model's own unembedding matrix. We apply this approach to \textit{VLMs} to interpret the model's understanding of \textit{visual information} in the model's textual vocabulary.

\subsection{Interpreting latent representations in Vision Models}
Understanding the internal dynamics of vision models is critical for ensuring safety and reliability in multimodal systems. Early works in this area focused on producing saliency maps~\citep{petsiuk2018rise}, analyzing individual neurons~\citep{bau2020units,bau2019gandissect,dravid2023rosetta}, and training networks to map latent representations to concepts~\citep{esser2020disentangling}. With the emergence of transformer-based vision models like CLIP~\citep{clippaper}, recent methods explain latent tokens ~\citep{chen2023interpretingvisionviatext} and the roles of attention heads and neurons with natural language~\citep{gandelsmanclipdecomposition,gandelsman2024interpretingsecondordereffectsneurons}. Few works currently interpret the internal computation of VLMs: \cite{palit2023visionlanguagemechinterp} develop a neuron causal tracing tool; \cite{schwettman2023multimodalneurons} identifies multi-modal neurons; and \cite{huo2024MMNeuron} ablates domain-specific neurons to improve vision question-answering. Whereas past papers have primarily studied the mechanisms (e.g. neuron analysis) that drive VLMs, we focus on interpreting and editing their latent representations for real-world applicability. 

\subsection{Detecting and reducing VLM hallucinations}

While VLM performances on image caption and visual question answering are continually improving, they continue to hallucinate facts that are not supported by the visual input. Existing methods for detecting hallucinations in language models during inference utilize latent representations~\citep{he2024llmfactoscopeuncoveringllms,su2024unsupervisedrealtimehallucinationdetection}, activations~\citep{chen2024insidellmsinternalstates}, and output logit values~\citep{varshney2023stitchtimesavesnine}. SAPLMA~\citep{azaria-mitchell-2023-internal} trains a hallucination classifier on the internal latent representations. LUNA~\citep{song2024lunamodelbaseduniversalanalysis} learns a transition function on latent representations and identifies abnormal transitions. \cite{varshney2023stitchtimesavesnine} uses the final layer logits to score the model’s confidence in an entity or keyword and intervenes by instructing the model to either repair or remove the hallucinated information. Among VLMs, LURE~\citep{zhou2024analyzingmitigatingobjecthallucination} is a fine-tuned revisor model to detect and reduce hallucinations. OPERA~\citep{huang2024operaalleviatinghallucinationmultimodal} uses the model's internal attention weights to detect and suppress patterns that align with the beginning of hallucinated phrases. In contrast to these methods, we leverage the internal \textit{image} representations in the VLMs for hallucination reduction and for zero-shot segmentation.


\section{Extracting Knowledge from VLMs}

We start by introducing VLMs and the general framework of their architectures in most recent work. We then describe our approach for decoding the features in intermediate image representations in VLMs into text, and apply it to two types of VLMs. Surprisingly, this approach effectively probes the knowledge about objects present in images and can localize objects within the image. 

\subsection{Preliminaries}
\label{subsec:preliminaries}

\textbf{Vision-Language Models.} The architecture of recent state-of-the-art VLMs for text generation typically involves three main components: a vision encoder to process image inputs, a mapping network to map image features to image embeddings, and an autoregressive language model to process the image embeddings and prompt embeddings to generate text. We focus on two recent state-of-the-art VLMs: \textit{LLaVA} 1.5 \citep{liu2024improvedbaselinesvisualinstruction} and \textit{InstructBLIP} \citep{dai2023instructblipgeneralpurposevisionlanguagemodels}. We use 7B versions of both these models. LLaVA utilizes a frozen CLIP vision encoder and an MLP as a mapping network to project the vision encoder outputs into image embeddings for the language model. The MLP is pre-trained on a large vision-language dataset and both the MLP and the language model are fine-tuned on an instruction-focused dataset. In contrast, InstructBLIP freezes both the vision encoder and the language model and only trains the mapping network. 


\textbf{Notations.} For the purposes of our work, we define the VLM architecture as follows. The vision encoder processes an input image to produce $n$ image features. These image features are projected to embedding space via the mapping network, resulting in $n$ $d$-dimensional image embeddings $\{k_i : k_i \in \mathbb{R}^d, i = 1, ..., n\}$. For the language model, the entire set of text tokens constitutes the vocabulary $V$ with vocabulary size $|V|$. The image embeddings, followed by $m$ text embeddings $\{t_i : t_i \in \mathbb{R}^d, i = 1, ..., m\}$ of the prompt tokens, are input to the language model through $L$ decoder layers. For an input embedding $x \in \mathbb{R}^{d}$, we define $h_l (x) \in \mathbb{R}^{d}$ to be the latent representation for embedding $x$ at layer $l \in \{1,...,L\}$, the output of the decoder layer, which is conditioned on previous tokens of the input sequence. An unembedding matrix $W_U \in \mathbb{R}^{|V| \times d}$ maps the last latent representation $h_L (t_{m})$ to a probability distribution over the vocabulary for the next token $t_{m+1}$.


\textbf{Logit Lens.} Logit Lens is an interpretability method for intermediate language model representations introduced in \Cref{sec:language_interpretability}. The logit lens technique applies the unembedding matrix $W_U$ to latent representations $h_l(x)$ in the $L$ intermediate layers in the language model to retrieve the logit distributions over the vocabulary.
\begin{equation}
f_l(t_{m}) = W_U \cdot h_l(t_m) = [\text{logit}_1, \text{logit}_2, \text{logit}_3, \ldots, 
\text{logit}_{|V|}]
\end{equation}
This is the logit distribution representing the predictions of the model after $l$ layers, where $\text{logit}_j$ corresponds to the token $j$ in the vocabulary. 

\subsection{Applying Logit Lens on VLMs}
\label{sec:logit_lens_vlms}

We apply the logit lens to probe the language model as it processes the image representations. This enables us to interpret the image features’ output distributions as they are transformed by the layers of the language model and localize objects spatially within the image.

\textbf{Extracting probability distributions from intermediate image representations}. We apply logit lens on the \textit{image representations} in the VLM. For a given image embedding $k_i$, we find the latent representation of the image embedding at layer $l$, $h_l(k_i)$, taking the logit lens to get the probability distribution over the vocabulary, $\text{softmax}(f_l(k_i))$. We define an object $o$, an object word composed of tokens from the vocabulary. We inspect the probability of a specific object $o$, $\text{softmax}(f_l(k_i))_o$. For multi-token objects, we take the maximum probability value over the object tokens. This provides a generalizable framework for analyzing specific latent image representations via text, with respect to specific objects. Next, we find the maximum probability over all image representations over all layers. For object $o$, we compute:

\begin{equation}
c_o = \max_{\substack{1 \leq l \leq L \\ 1 \leq i \leq n}} \{\text{softmax}(f_l(k_i))_o\}
\end{equation}

We define $c_o$ as the VLMs \textit{internal confidence} of an object $o$ existing in the image: the highest probability of object presence across $n$ image representations through $L$ layers of the language model.

\textbf{Comparing the internal confidence of present and not present objects.} To determine if internal confidence provides meaningful information about objects in the image, we examine $c_o$ for objects present and not present in an image. We use InstructBLIP and LLaVA to caption 5000 random COCO2014 images in the Karpathy validation split~\citep{lin2015microsoftcococommonobjects} and determine $c_o$ for all 80 COCO objects, only a few of which are present in each image. Since there are many more objects not present than present, we randomly sample a subset of the internal confidences for objects not present. Figure~\ref{figure:logit_and_softmax} exhibits the internal confidences for objects present and not present in the image. We empirically find that the VLMs' internal confidences are higher for present objects than not present ones. We use this claim later to classify objects as hallucinations in~\Cref{sec:hallucination_detection}.

\begin{figure}[t!]
    \centering
    \includegraphics[width=\textwidth]{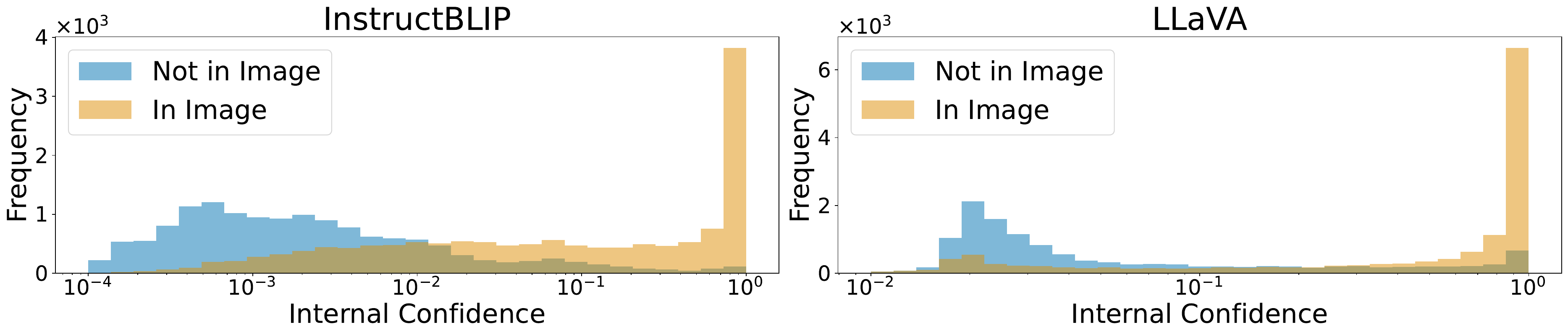}
    \vspace{-1em}
    \caption{\textbf{Comparison of internal confidence in objects present and not present in the image}. We examine the internal confidence of COCO objects that exist and do not exist in the image within intermediate VLM image representations. We observe that objects that do not exist in the image have lower internal confidence.}
    \label{figure:logit_and_softmax}
\end{figure}
\begin{figure}[t!]
    \centering
    \includegraphics[width=\textwidth]{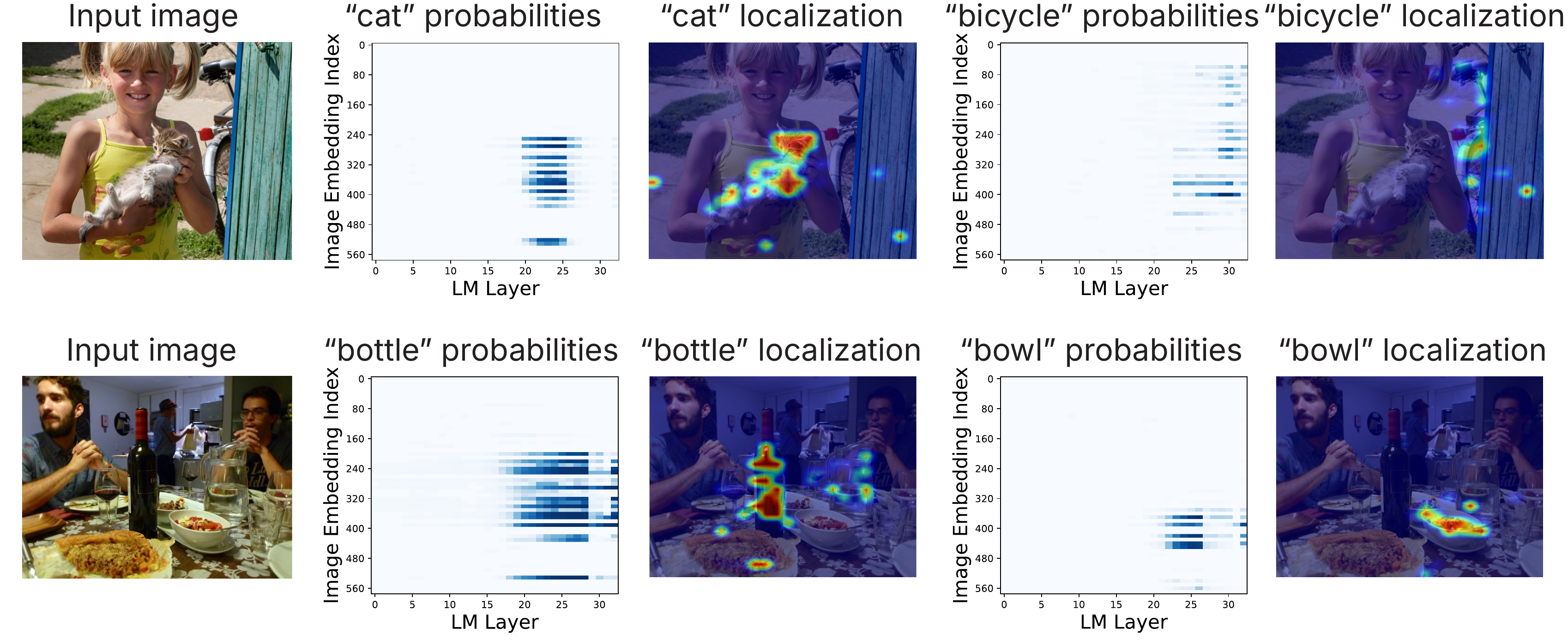}
    
    \vspace{-1em}\caption{
    \textbf{Localizing objects using internal confidence values}. We find the probabilities of objects through layers of the language model for every image embedding in LLaVA. We use the highest layer probability per image embedding to localize an object within the image.
    \label{figure:visual_grounding}}
\end{figure}

\textbf{Object localization}. Given that the language model can distinguish between objects present and not present in an image, we examine whether it can attribute high object internal confidence to specific patches in an image. For each image embedding $k_i$ in $n$ image embeddings, we find the maximum softmax probability of an object within the layers of the model, $\max_{1 \leq l \leq L} \{\text{softmax}(f_l(k_i))_o\}$. Using these internal confidence values, we localize the objects in the image patches, each of which maps to an image embedding. We focus on LLaVA for this task, since its image encoder preserves the spatial mapping of image patches to image features.

We observe that image representations that exhibit higher internal confidence for specific objects correspond to the image patches in which those objects are visually present (examples in \Cref{figure:visual_grounding}). Building on our previous observation, we see that the intermediate image representations semantically align with latent token representations of objects present in them while maintaining their spatial locality. We use this unique finding for zero-shot segmentation in \Cref{sec:segmentation}.

While the model is not directly trained to map the image representations closer to the text representations of objects within them, we can unembed the image representations in the text vocabulary for localization and find differences in internal confidence for present and hallucinated objects. In \Cref{sec:hallucination_detection}, we will use this observation for various applications including hallucination detection and zero-short segmentation.

\section{Erasing knowledge from VLMs}

Recognizing that image embeddings are directly interpretable (\Cref{sec:logit_lens_vlms}), we edit these embeddings to erase the presence of objects from image captions. We propose a linear editing algorithm that subtracts the text embedding of a target object from all image embeddings. When applied on singular and multiple object removals, we find that it erases hallucinated objects more effectively than correctly detected (CD) objects (i.e. real objects that the model correctly detects). 

\subsection{Erasing objects from image representations}
\label{sec:sub:individual_objects}

\begin{wrapfigure}{r}{0.5\textwidth}
\vspace{-2.5em}
\begin{minipage}{\linewidth}
\begin{algorithm}[H]
\caption{\textsc{ProjectAway}}
\begin{algorithmic}
\State \textbf{Input:}  A set of image embeddings $K$, text embedding $\vec{t}$, and weight factor $\alpha$
\State \textbf{Output:} A set of modified image embeddings $K'$ projected away from the text embedding
\State \textbf{Initialization:} $K' \gets \emptyset$
\For{$\vec{k} \in K$}
\State $p \gets \vec{k} \cdot \vec{t}$
\If{$p > 0$}
    \State $K' \gets K' \cup \{ \vec{k} - \alpha \cdot \frac{p}{\lVert \vec{t} \rVert_2^2} \cdot \vec{t} \}$
\Else
    \State $K' \gets K' \cup \{\vec{k} \}$
\EndIf
\EndFor
\end{algorithmic}
\end{algorithm}
\vspace{-1.5em}
\caption{Our editing algorithm erases the presence of an object from image embeddings by orthogonalizing them with respect to the object's text embedding.}
\label{algorithm:project_away}
\end{minipage}
\end{wrapfigure} 

We present an algorithm, \textsc{ProjectAway} (\Cref{algorithm:project_away}), that orthogonalizes image representations with respect to text representations in order to erase objects in image captions, applying it to remove objects one at a time and all at once.

Given an image and an object to remove, we edit the latent representations $h_{l^{I}}(k_i)$ at a hidden layer $l^{I}$ across all image embeddings $k_i$. We do not modify any latent representations outside of those belonging to image features. We compute the dot product, $p$, of $h_{l^I}(k_i)$ and the object's text embedding $\vec{t}$, subtracting a weighted $\vec{t}$ from $h_{l^I}(k_i)$ only if the dot product is positive. At $\alpha=1$, $\textsc{ProjectAway}$ is equivalent to orthogonalizing the image representations with respect to the text representation. To compute text representation $\vec{t}$, we pass the object (e.g. ``hot dog'') into the VLM's text model and extract $h_{l^{T}}(t_{\text{-1}})$ at hidden layer $l^{T}$, where $t_{\text{-1}}$ is the last token of the object. We use the last token of the object to capture the whole of the object's meaning. 

\subsubsection{Removing objects one by one}
\label{sec:sub:individual_removal}

We evaluate the \textsc{ProjectAway} algorithm's effectiveness at erasing individual objects from captions across multiple images and objects.


\textbf{Experimental setting.} We apply \textsc{ProjectAway} on 5000 random images from the COCO2014 training set on all mentioned COCO objects (i.e. hallucination and CD) individually and measure the removal rate at which objects no longer appear in the caption. For InstructBLIP, we set $(l^{I}, l^{T}, \alpha) = (1, 2, 1.5)$. For LLaVA, we set $(l^{I}, l^{T}, \alpha) = (19, 21, 3.5)$. These parameters are fixed irrespective of image and are chosen for their maximal effect (see ablations in \Cref{sec:sub:ablations_mass_removal}). To differentiate hallucinations from CD, we compute CHAIR~\citep{rohrbach2019objecthallucinationimagecaptioning}, an evaluation criteria that compares model-generated captions to ground-truth human annotations. CHAIR provides two main scores, $\text{CHAIR}_I$ and $\text{CHAIR}_S$, that quantify hallucinations for instances and sentences, respectively:
\begin{equation}
\text{CHAIR}_S = \frac{|\{\text{captions with hallucinated objects}\}|}{|\{\text{all captions}\}|}, \text{CHAIR}_I = \frac{|\{\text{hallucinated objects}\}|}{|\{\text{all objects mentioned}\}|}
\end{equation}

\textbf{Results.} \Cref{table:mass_removing} shows that \textsc{ProjectAway} is significantly more effective in erasing individual hallucinated objects at an individual level than CD objects for both InstructBLIP and LLaVA. Along with the insight that hallucinated objects have lower softmax scores (\Cref{figure:logit_and_softmax}), these results suggest that hallucinated objects manifest more weakly in image embeddings and are hence easier to remove than CD objects.



\subsubsection{Mass-removing objects}
\label{sec:sub:mass_editing}
We iteratively apply \textsc{ProjectAway} to a \textit{set} of objects, following the same experimental setup and observing similarly different removal rates for hallucinated objects and CD objects.

\begin{figure}[t!]
    \centering
    \includegraphics[width=\textwidth]{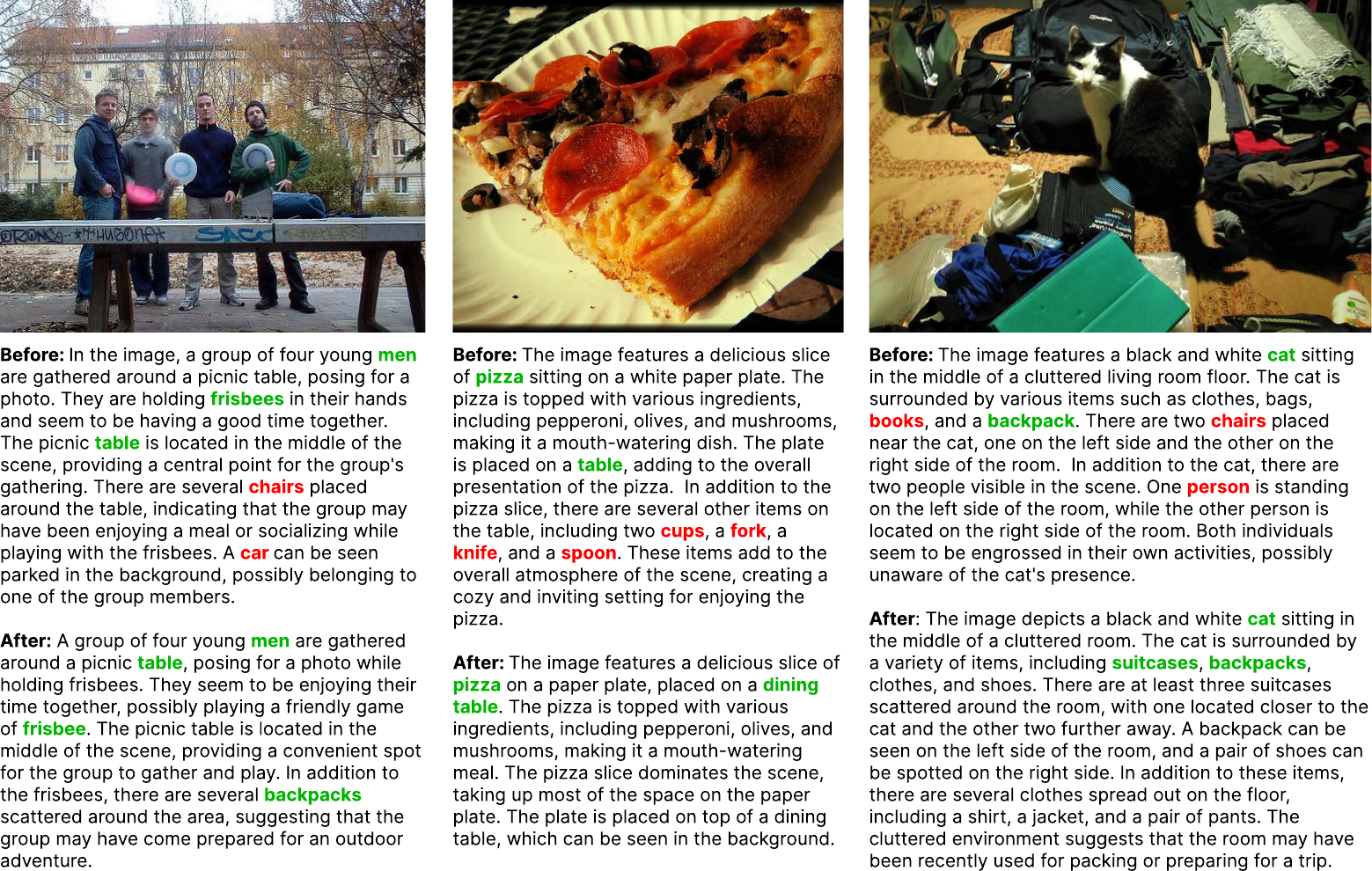}
    
    \vspace{-1em}\caption{\textbf{Qualitative results for mass object removal.} We present example images and their captions after mass-removing hallucinations (\textcolor{red}{red}) with \textsc{ProjectAway.}, which can effectively remove hallucinations while preserving, even increasing, correctly detected objects (\textcolor{green}{green}).
    \label{figure:hallucination_examples}}.
    
\end{figure}

\textbf{Mass-removing hallucinations.} We mass-remove hallucinations identified with ground truth annotations using $\textsc{ProjectAway}$. \Cref{table:mass_removing} shows that editing out all the hallucinations of an image yields a similar removal rate as individually editing out and, importantly, that erasing hallucinated objects together does not interfere with each other. We achieve a hallucination reduction rate of 41.3\% for InstructBLIP and 23.3\% for LLaVA (see \Cref{table:mass_removing_appendix}). Recall count slightly \textit{increases} for both models, indicating that caption accuracy is preserved. This may be because removed hallucinations are replaced with objects the model is more confident in. Qualitative results are in \Cref{figure:hallucination_examples}.

\begin{table}[t]
\small
\caption{\textbf{Removing mentioned objects individually \& in-mass.} Using $\textsc{ProjectAway}$, we remove hallucinated objects and observe high hallucination reduction with CHAIR, mass-removal rate (Mass RR), and individual removal rate (Individual RR). We also remove correctly detected (CD) objects but find that they are more resistant to linear editing. Denote CHAIR$_S$ as $C_S$ and CHAIR$_I$ as $C_I$.}
\begin{center}
\small
\begin{tabular}{lllllll} 
\toprule
\bf Edit Scope & \bf Model & \bf Individual RR (\%) & \bf Mass RR (\%) & \bf CD change (\%) & \bf C$_i$ $\downarrow$ & \bf C$_s$ $\downarrow$   
\\
\midrule 
\multirow{2}{*}{No edits} & InstructBLIP & - & - & - & 15.0 & 54.1 \\ 
& LLaVA & - & - & - & 14.6 & 51.1  \\ 
\hline 
\addlinespace
\multirow{2}{*}{Hallucinations} & InstructBLIP & 83.3 & 74.3 & +0.07 & 8.94 & 33.2  \\ 
& LLaVA & 86.0 & 72.8 & +0.01 & 11.2 & 35.5  \\ 
\hline 
\addlinespace
\multirow{2}{*}{CD} & InstructBLIP & 16.2 & 15.0 & -2.2 & 17.3 & 58.3  \\ 
& LLaVA & 6.9 & 8.3 & -1.6 & 15.2 & 52.4  \\ 
\bottomrule
\end{tabular}
\label{table:mass_removing}
\end{center}
\end{table}




\textbf{Mass removing CD.} We similarly find that applying \textsc{ProjectAway} can successfully remove CD objects when edited all together in \Cref{table:mass_removing}. Furthermore, CHAIR scores minimally change, which indicates that this mass-removal merely erases object presence without eroding caption accuracy. While the removal rate is lower than for hallucinated objects, this insight proves useful when we apply \textsc{ProjectAway} for hallucination reduction in \Cref{sec:hallucination_reduction}.





\subsection{Ablation Study: mass-removing hallucinations}
\label{sec:sub:ablations_mass_removal}

We perform ablations on parameters of \textsc{ProjectAway} to improve object removal rate for erasing hallucinations in-mass. 

\textbf{Experimental setting.} We ablate the three parameters of \textsc{ProjectAway}: layer $l^I$ to edit at, layer $l^T$ to retrieve the text representation, and weight factor $\alpha$. At $l^T = -1$, we average together the object's constituent token embeddings. At $l^I = -1$, we edit the image embeddings directly inputted to the text model. We evaluate across 500 training samples from COCO 2014 that have at least one hallucination. 

\begin{figure}
\centering
    \begin{floatrow}
\ffigbox{%
  \includegraphics[width=0.50\textwidth]{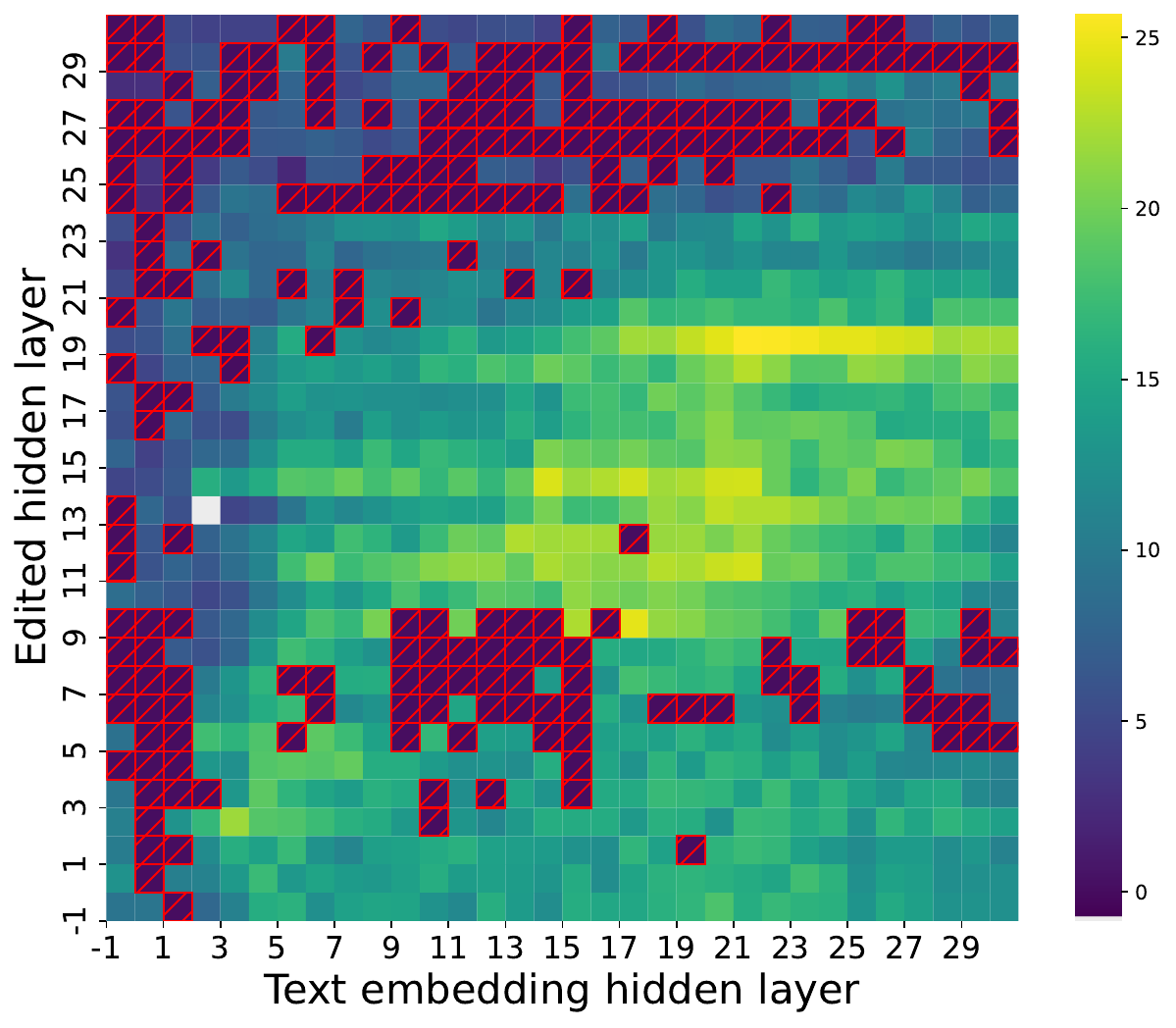}
  }{
  \vspace{-1.7em}
\caption{
\textbf{Hidden layer ablations} for LLaVA. We track hallucination reduction (\%) across different layers to edit at and extract latent embeddings for the text embedding, crossing out (red) parameters from consideration where there is a decrease in correctly detected objects.
\label{fig:hidden_layer_ablations}
}}%
\ffigbox{%
\centering
  \includegraphics[width=0.50\textwidth]{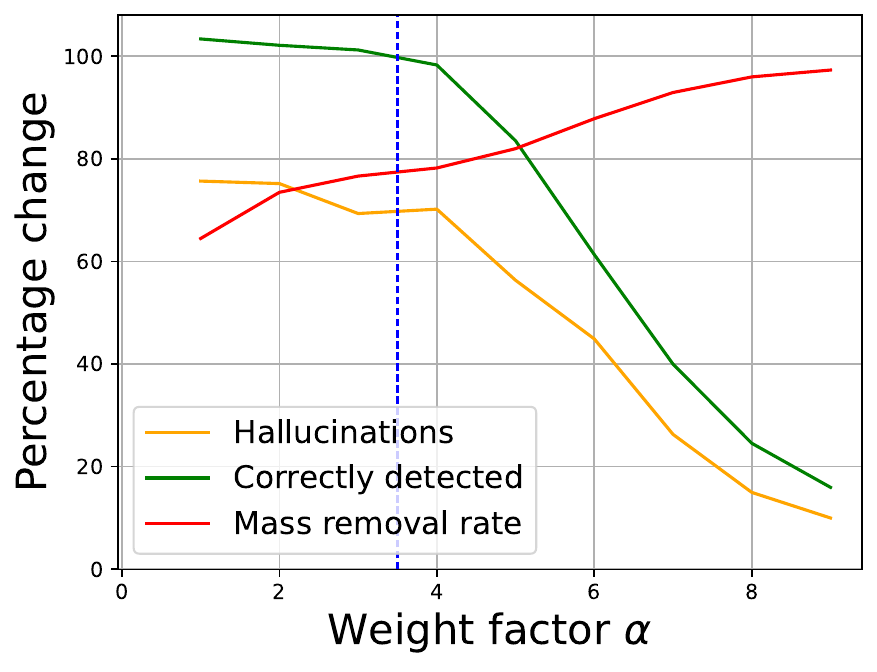}
  }{
  \vspace{-1em}
\caption{
\textbf{Weight ablations} for LLaVA. We vary the weight factor $\alpha$ and measure changes in correctly detected objects, removal rate, and hallucination reduction. We observe a decline in hallucinations as weight grows and mark a weight where there is no loss in correctly detected objects.
\label{fig:weight_ablations}
}}
\end{floatrow}
\end{figure}

\label{Section 4.2} 
\textbf{Hidden layers.} \Cref{fig:hidden_layer_ablations} shows hallucination reduction rate on LLaVA from mass-removing hallucinations on every combination of $l^I$ and $l^T$ (each from -1 to 31). As a core concern is that editing erodes caption accuracy, we gray out any combination that reduces CD objects. For InstructBLIP (see \Cref{fig:hidden_layer_ablations_blip}), the best parameters $(l^I = 1, l^T = 2)$ reduces hallucinations by 38.5\%. For LLaVA, our best parameters $(l^I = 19, l^T = 21)$ reduce hallucinations by 25.7\%, and the middle layers are the best to edit and extract latent text embeddings from. Our results also provide a wide range of reasonable parameter alternatives to use if this reduction rate does not generalize beyond our samples. 

\textbf{Weight factor.} Using the best-reduced hidden layers, we ablate the weight factor $\alpha$ for \textsc{ProjectAway} across the same 500 randomly selected COCO images. \Cref{fig:weight_ablations} shows that as $\alpha$ increases, hallucinations are removed at a higher rate, and the overall hallucination count drops significantly. At high $\alpha$, we observe through anecdotal examples that captions become nonsensical, as quantitatively shown by the complete loss of both correctly detected and hallucinated objects from the caption. 
Therefore, as a pre-caution, we only select weight factors that do not reduce CD objects when we apply \textsc{ProjectAway} to erase hallucinated objects.

\section{Applications}

\subsection{Hallucination Detection}
\label{sec:hallucination_detection}
When extracting knowledge from VLMs in \Cref{sec:logit_lens_vlms}, we found that applying logit lens on in-context image representations exhibit useful information about visual objects present in the image. Using these observations, we present an approach for object presence classification that only relies on the VLMs own parameters. We utilize the internal confidence $c_o$ value to classify object presence, since the internal confidence for objects that are not present in the image, or hallucinated, are lower within the image representations.

\textbf{Experimental setting.} We evaluate the strength of the internal confidence $c_o$ as an indicator of object presence. We sample 5000 images from the MSCOCO training set, using the image captioning objective to caption methods with both InstructBLIP and LLaVA. We use the $c_o$ for present objects and hallucinations within the captions generated by each VLM. We assess how well the internal confidence aligns with the ground truth labels of object presence, where a negative sample is a hallucination and a positive sample is a present object.

\textbf{Baseline.} As a baseline, we use the maximum output probability of the object's tokens. This is the confidence of the model prediction. Previous works such as \cite{zhou2024analyzingmitigatingobjecthallucination} have found that hallucinations occur more frequently on objects characterized by high uncertainty during generation.

\begin{figure}[t]
    \centering
    \includegraphics[width=\textwidth]{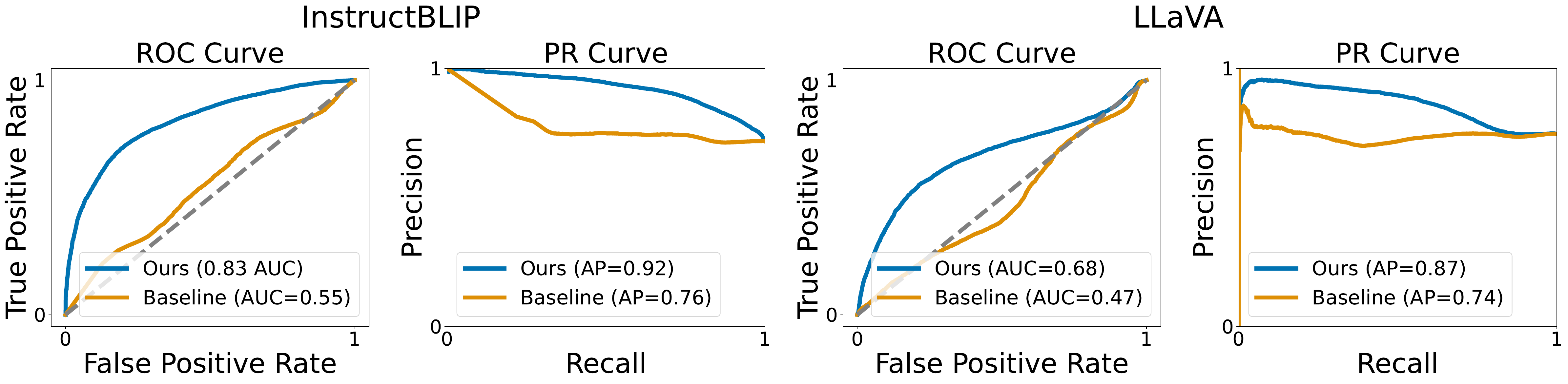}
    \vspace{-1.5em}
    \caption{\textbf{Object Presence Classification Curves for InstructBLIP and LLaVA.} We show the Precision-Recall and ROC curves of our confidence measure for present object-hallucination classification on the COCO training subset. Classifying object presence with the internal confidence outperforms the baseline, indicating that the language model's image representations know which objects are hallucinations and which are truly present.}
    \label{fig:hallucination_detection}
\end{figure}

\textbf{Results.} We present quantitative results in \Cref{fig:hallucination_detection} and \Cref{table:hallucination_detection}. We show qualitative results for LLaVA (\Cref{figure:llava_object_presence}) and InstructBLIP (\Cref{figure:blip_object_presence}) in the Appendix. We find that utilizing internal confidence to classify object hallucinations provides a 47.17\% improvement in mAP in InstructBLIP and 22.45\% in LLaVA. Furthermore, the ROC AUC improves over the baseline by 50.10\% in InstructBLIP and 44.68\% in LLaVA, indicating stronger object presence classification. 

\subsection{Hallucination Removal}
\label{sec:hallucination_reduction}

We use the mass editing technique to remove hallucinations detected by the prior method. \Cref{sec:sub:mass_editing} successfully removes a significant portion of hallucinations but presupposes a knowledge of what the hallucinations are. We threshold on the internal confidence of each object to identify hallucinations and mass-remove them using \textsc{ProjectAway}. Our chosen threshold prioritizes precision over recall (i.e. we allow classification of some CD objects as hallucinations) because CD objects are less affected by the removal method, as shown in \Cref{sec:sub:mass_editing}.

\textbf{Experimental setting.} We threshold hallucinations as $c_o < 0.2$ for InstructBLIP and $c_o < 0.1$ for LLaVA. Based on prior ablations (\Cref{Section 4.2}), we select $(l^I = 1, l^T = 2, \alpha = 1.5)$ for InstructBLIP and $(l^I = 19, l^T = 21, \alpha = 3.5)$ for LLaVA. Our prompt is ``Please describe this image in detail." 

\textbf{Baselines.} Since our method intervenes during the decoder step, we compare our method with 3 standard decoding algorithms. Greedy decoding predicts the next token based on the highest logit probability. Beam search maintains a tree of beams and selects the best beam at generation end. Nucleus sampling selects the next token from a set of high probability tokens whose cumulative probability reaches a threshold $p$. We also evaluate against OPERA~\citep{huang2024operaalleviatinghallucinationmultimodal}, which mitigates hallucinations by adding an overtrust penalty during decoder generation. We set $p = 0.9$ for nucleus sampling. We use beam search in our method and unify $N_{\text{beam}} = 5$ for the baseline.

\textbf{Results.} We apply these parameters to 500 COCO images from the Karpathy validation set. We provide qualitative results in \Cref{figure:blip7b_hallucination_reduction_samples} and \Cref{figure:llava7b_hallucination_reduction_samples}. Quantitative results in \Cref{table:hallucination_removal} show that we outperform our baselines and reduce hallucinations by 25.7\% on InstructBLIP and 23.8\% on LLaVA compared to beam search. Our approach achieves a similar hallucination reduction rate as \Cref{sec:sub:mass_editing}, despite not precisely differentiating hallucinations and some CD objects being incorrectly edited out. Notably, our method relies on no training or external models, effectively offering a ``free lunch." We find similar performance on additional models (\Cref{appendix:more_advanced_models}) and attribute hallucinations (\Cref{appendix:attribute_hallucinations}).

\begin{table}[t]
\small
\caption{\textbf{Hallucination intervention performance.} We mass-remove hallucinations detected by the method in \Cref{sec:hallucination_detection} and outperform other baselines. We observe a considerable drop in the raw count of hallucinated objects.}
\begin{center}
\label{table:hallucination_removal}
\begin{tabular}{lllll} 
\toprule
\bf Model & \bf Method & \bf CHAIR$_i \downarrow$ & \bf CHAIR$_s \downarrow$ & \bf Hallucinated Objects $\downarrow$
\\
\midrule 
\multirow{5}{*}{InstructBLIP} & Greedy & 57.0 & 23.3 & 512 \\
& Nucleus & 58.0 & 24.0 & 508\\
& Beam Search & 53.4 & 14.6 & 564  \\
& OPERA & 45.6 & 13.9 & 472 \\ 
& Ours & \textbf{43.8} & \textbf{12.5} & \textbf{419} \\
\midrule
\multirow{5}{*}{LLaVA} & Greedy & 49.2 & 14.2 & 532\\
& Nucleus & 55.8 & 17.1 & 618\\
& Beam Search & 52.4 & 15.0 & 583\\
& OPERA & 44.8 & 12.8 & 462  \\ 
& Ours & \textbf{42.0} & \textbf{12.2} & \textbf{444} \\
\bottomrule
\end{tabular}
\end{center}
\end{table}


\subsection{Zero-shot Segmentation}
\label{sec:segmentation}

Building upon our findings in \Cref{sec:logit_lens_vlms}, we utilize the internal confidence per image feature for zero-shot image segmentation. This application leverages the spatial information encoded in the image representations and demonstrates how VLMs internally represent and localize objects within images.

\textbf{Method.} Our approach leverages the spatial correspondence between image patches and their associated image embeddings. We use LLaVA to generate the name of the class in the image and we focus on the internal confidence of that class per image patch. We take the mean internal confidence for tokens comprising a class word. We resize the set of $24\times24$ internal confidence values per image patch back into a fixed image size of $336\times366$ pixels. We then apply a threshold to these confidence values to binarize them into a foreground/background segmentation for the object in the image.

\begin{wrapfigure}{r}{0.5\textwidth}
    \centering
    \includegraphics[width=\linewidth,trim={0 0 0 2.5em},clip]{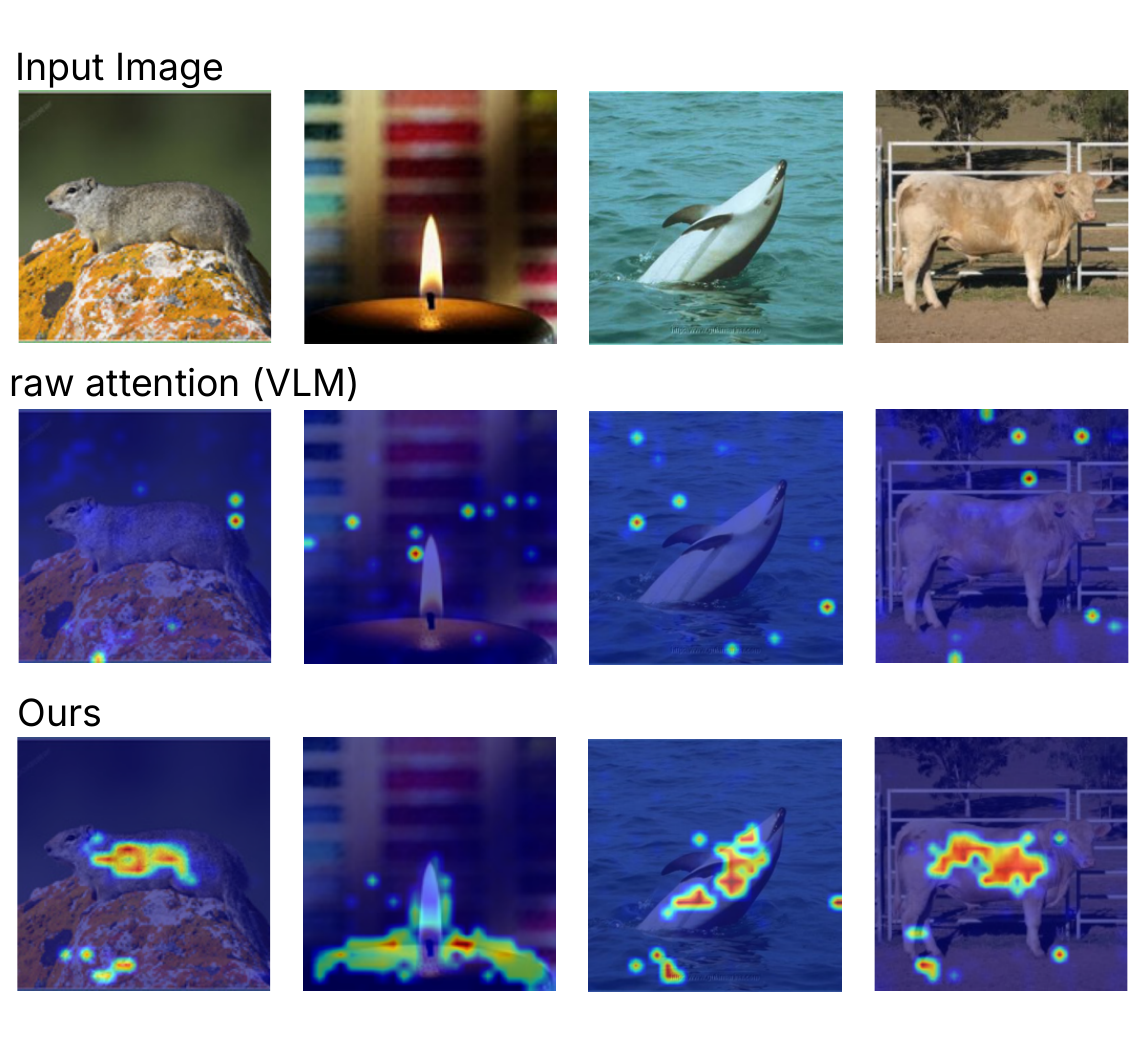}
    \vspace{-2.5em}
    \caption{\textbf{Zero-shot segmentation.} Warmer areas indicate higher internal confidence for the class at that image patch. We binarize these values with a threshold to generate segmentations.}
    \label{figure:imagenet_segmentation}
\end{wrapfigure}

\textbf{Baseline.} As a baseline, we extract the attention values of generated tokens with the image embeddings from LLaVA. We also compare to the segmentation method introduced by~\cite{gandelsmanclipdecomposition}, which utilizes the attention heads of the image encoder without the additional VLM processing, using the same image encoder (CLIP-ViT-L/14 at 336px).

\textbf{Results.} We evaluate our method on the Imagenet validation set. Qualitative results are shown in \Cref{figure:imagenet_segmentation} and quantitative comparisons with the baselines in \Cref{tab:segmentation_performance}. We improve mAP by 8.03\% over using the VLMs raw attention values and provide better and/or comparable performance to other state-of-the-art methods that utilize just the image encoder. While the VLM is not directly trained for segmentation, our technique reveals that it still encodes significant \textit{spatial} information about objects within its intermediate image representations. 

\begin{table}[t]
\small\begin{center}
\begin{tabular}{lcccc}
\toprule

\bf Model & \bf Method & \bf Pixel Acc. $\uparrow$ & \bf mIoU $\uparrow$ & \bf mAP $\uparrow$ \\
\midrule
raw attention (CLIP) & Image Encoder & 69.81 & 45.19 & 77.30\\
TextSpan~\citep{gandelsmanclipdecomposition} & Image Encoder & \underline{75.57} & \underline{53.60} & \textbf{80.22}\\
raw attention (VLM) & VLM & 67.28 & 39.27 & 73.96 \\
Ours & VLM & \textbf{76.16} & \textbf{54.26} & \underline{79.90} \\
\bottomrule
\end{tabular}
\caption{\textbf{Segmentation Performance on ImageNet-segmentation.} Localizing objects using their probabilities within the image representations results in more accurate zero-shot segmentation than previous methods relying on vision encoders and VLMs.}
\end{center}
\label{tab:segmentation_performance}
\end{table}

\section{Discussion and limitations}


We interpreted VLMs' image representations through the language model layers and discovered that linear editing of these representations can selectively remove object information via a simple orthogonalization. Our findings enabled hallucination reduction and improved zero-shot segmentation. We present two limitations of our work and conclude with future directions.

\textbf{Multi-token objects.} Our method simplifies the use of object words that may be composed of multiple tokens, such as by taking the max internal confidence over object tokens or utilizing the average token embedding for editing. This can introduce noise to the internal confidence if certain tokens are common in multiple different words and lead to an approximation of the object's latent representations when editing.

\textbf{Fine-grained edits.} The editing approach may struggle with highly abstract or longer sentences that involve attributes or interactions of objects. Removing a full sentence, for example, is not something we assessed in this paper, since our focus is on the removal of individual objects.

\textbf{Future work.} While our focus was on interpreting objects and object hallucinations in VLMs, we believe that our approach can be extended to other key elements of visual scenes, such as people, attributes, and actions. We also focused on object removal, but we believe that editing can also be extended to inject objects into a caption (by adding instead of subtracting the text embedding). We hope to explore the applications of our approach in other multimodal architectures. Our insights may help design better VLMs that are more robust to hallucinations and have improved spatial understanding. We plan to explore these directions in our future work.

\newpage

\subsection{Acknowledgments}

We thank Kayo Yin for her comments and feedback on our paper. YG is supported by the Google Fellowship. Authors, as part of their affiliation with UC Berkeley, were supported in part by the the Berkeley Artificial Intelligence Research (BAIR) commons program.

\bibliography{iclr2025_conference}
\bibliographystyle{iclr2025_conference}
\newpage

\appendix
\section{Appendix}
\subsection{Mass-removing objects}
\label{appendix:mass_removing}

We mass-remove mentioned objects (hallucinations and correctly detected) with \textsc{ProjectAway} and tally up the total number of unique hallucinated and CD objects in \Cref{table:mass_removing_appendix}.

\begin{table}[t]
\small
\caption{\textbf{Supplemental metrics for \Cref{table:mass_removing}.} We measure unique hallucinated and correctly detected (CD) objects.}
\begin{center}
\small
\begin{tabular}{lllllll} 
\toprule
\bf Edit Scope & \bf Model & \bf Hallucinations & \bf CD   
\\
\midrule 
\multirow{2}{*}{No edits} & InstructBLIP & 4545 & 14178 \\ 
& LLaVA & 4372 & 15053  \\ 
\hline 
\addlinespace
\multirow{2}{*}{Hallucinations} & InstructBLIP & 2672 & 14189 \\ 
& LLaVA & 3348 & 15061 \\ 
\hline 
\addlinespace
\multirow{2}{*}{CD} & InstructBLIP & 5078 & 13864  \\ 
& LLaVA & 4583 & 14826  \\ 
\bottomrule
\end{tabular}
\label{table:mass_removing_appendix}
\end{center}
\end{table}

\subsection{Ablations for InstructBLIP}

We show hidden layer and weight ablations for mass-removing hallucinations in InstructBLIP referenced in \Cref{Section 4.2}. The hidden layer ablations indicate that most of the parameter space is too sensitive to edit and leads to losses in correctly detected objects. We find that smaller $l^T$ and $l^I$ parameters are the most effective for reducing hallucinations. Our best parameters $(l^I = 1, l^T=2)$ reduce hallucinations by 38.5\%. It is not fully understood why the majority of the parameter search space is invalid in comparison with LLaVA in \Cref{fig:hidden_layer_ablations}. It is possible that the fine-tuning step in LLaVA semantically aligns hidden image representations with text embeddings more than InstructBLIP, allowing linear edits to have the precise, intended effect.

\begin{figure}
\centering
    \begin{floatrow}
\ffigbox{%
  \includegraphics[width=0.50\textwidth]{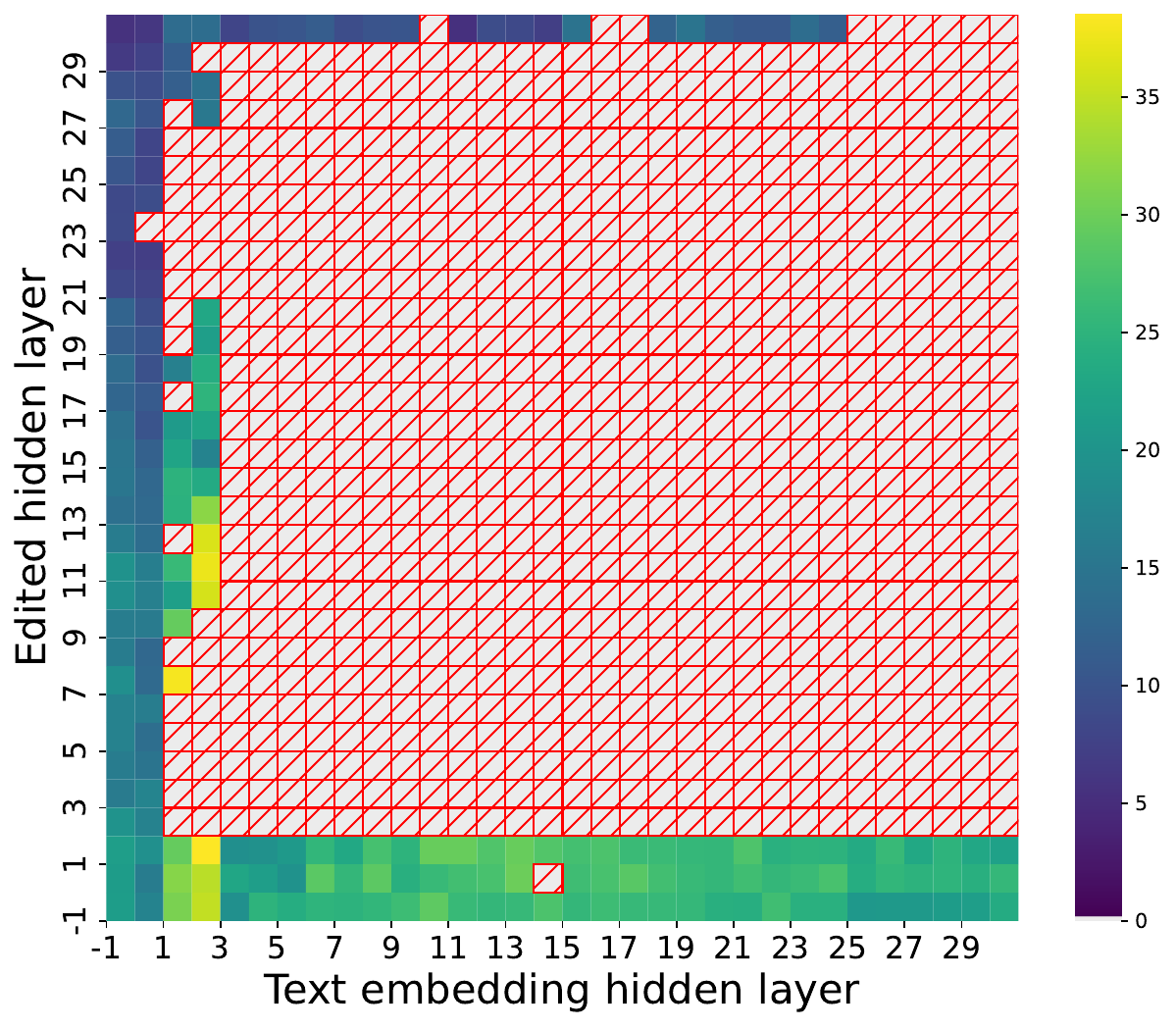}
  }{
  \vspace{-1.7em}
\caption{
\textbf{Hidden layer ablations for InstructBLIP}. We track hallucination reduction (\%) across different layers to edit at and extract latent embeddings for the text embedding, crossing out (red) parameters from consideration where there is a decrease in correctly detected objects.
\label{fig:hidden_layer_ablations_blip}
}}%
\ffigbox{%
\centering
  \includegraphics[width=0.50\textwidth]{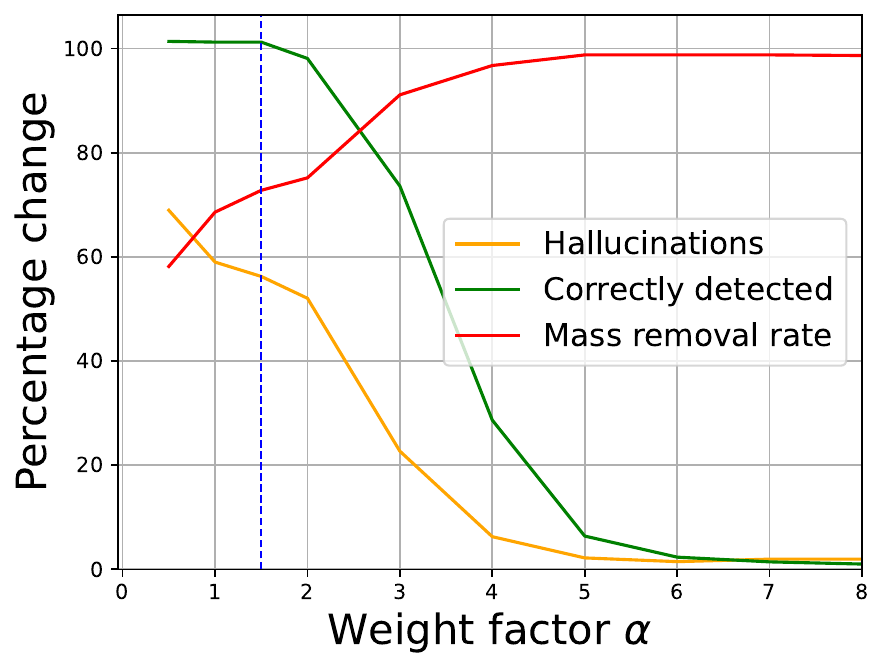}
  }{
  \vspace{-1em}
\caption{
\textbf{Weight ablations for InstructBLIP.} We vary the weight factor $\alpha$ and measure changes in correctly detected objects, object removal rate, and hallucination reduction. We observe a decline in hallucinations as weight increases and mark a weight where there is no loss in correctly detected objects.
\label{fig:weight_ablations_blip}
}}
\end{floatrow}
\end{figure}


\subsection{Hallucination Detection}

\begin{figure}[t!]
    \centering
    \includegraphics[width=\textwidth]{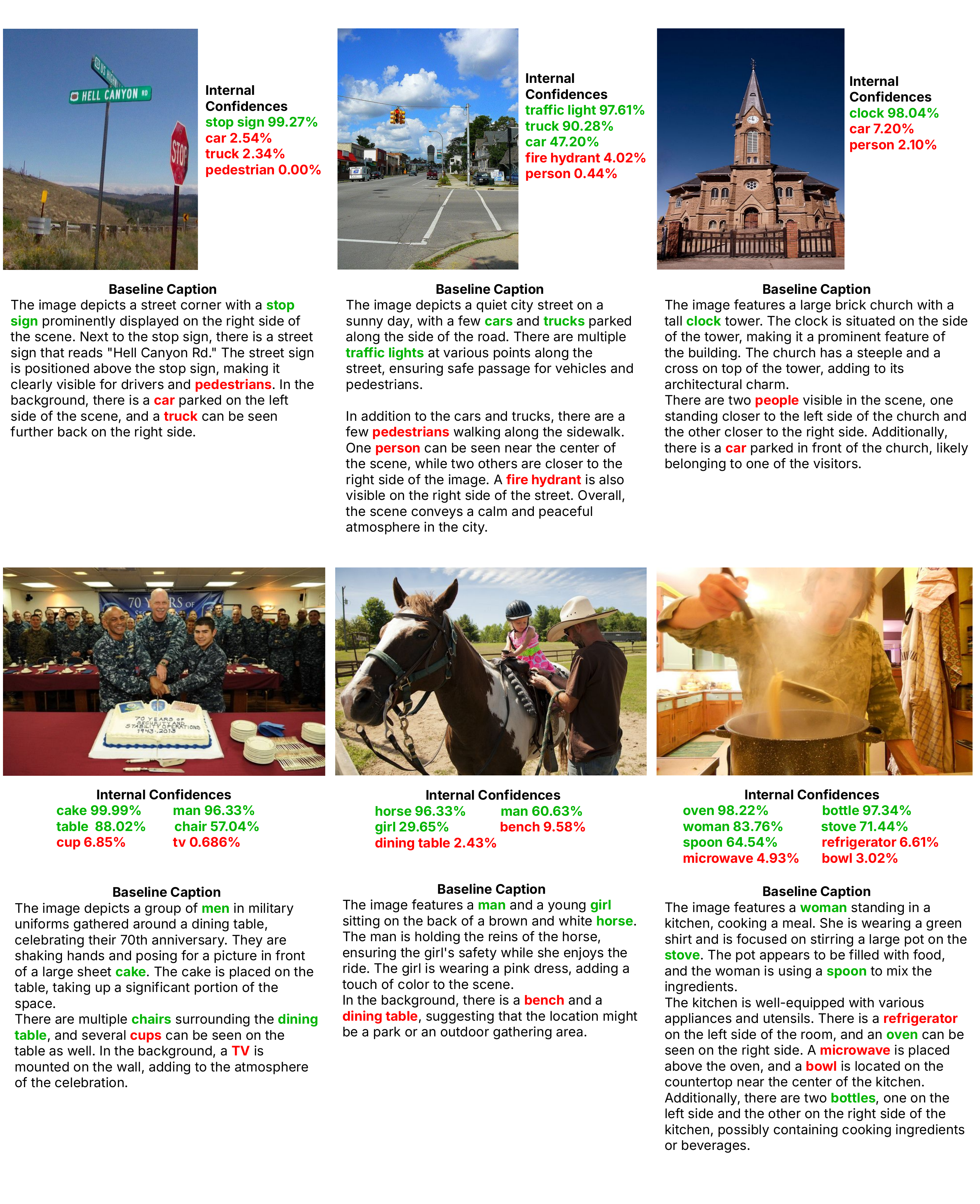}
    \vspace{-1em}
    \caption{\textbf{LLaVA Object Presence Classification}. Sample image captions from LLaVA and the internal confidence scores for objects in the caption used for classification as correctly detected objects or hallucinations.}
    \label{figure:llava_object_presence}
\end{figure}

\begin{figure}[t!]
    \centering
    \includegraphics[width=\textwidth]{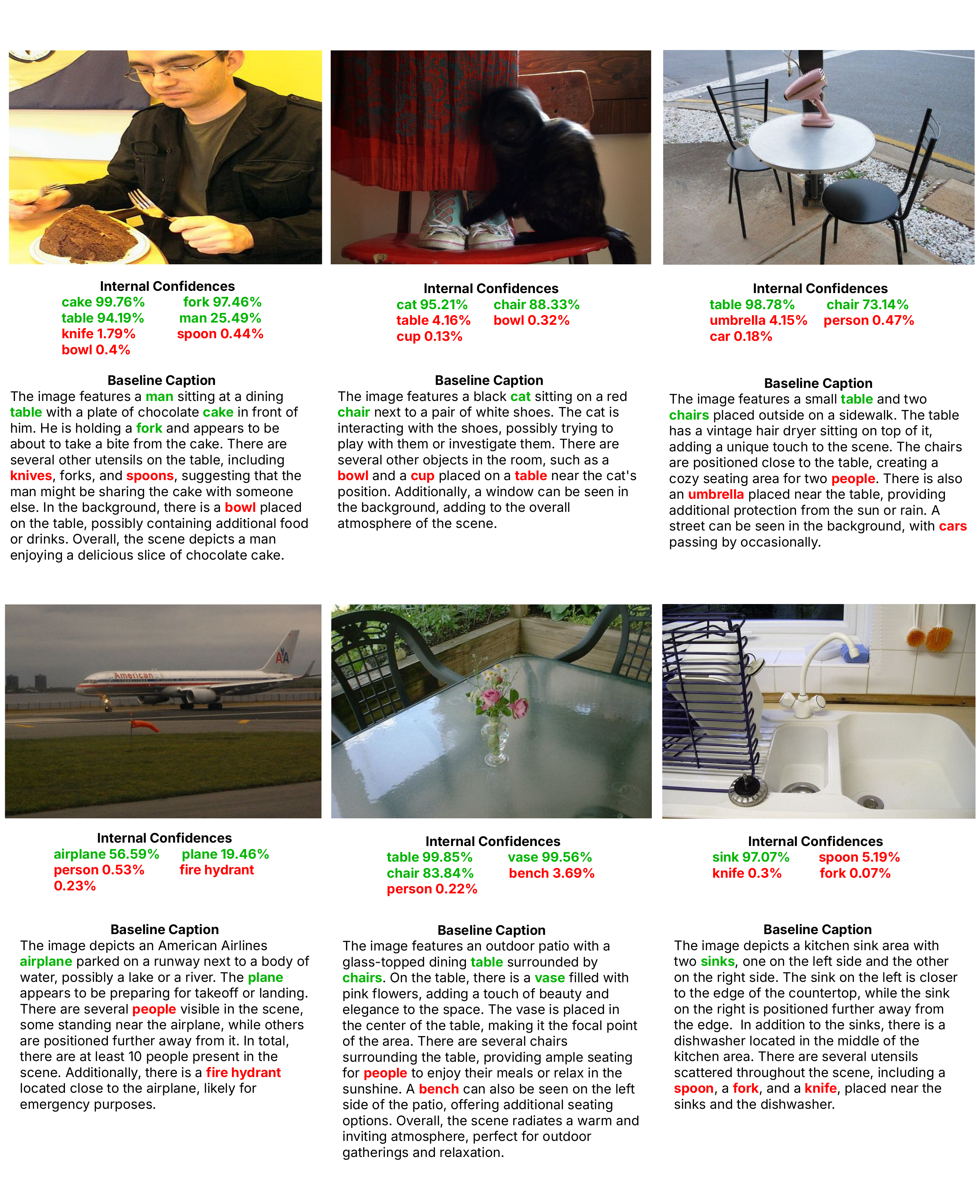}
    \vspace{-1em}
    \caption{\textbf{InstructBLIP Object Presence Classification}.}
    \label{figure:blip_object_presence}
\end{figure}

We show quantitative comparisons from our hallucination detection approach using internal confidence (\Cref{sec:hallucination_detection}) to the baseline in \Cref{table:hallucination_detection}. We also show qualitative examples for LLaVA in \Cref{figure:llava_object_presence} and for InstructBLIP in \Cref{figure:blip_object_presence}. These samples exhibit model-generated captions, parsed objects, and whether they are classified as hallucinated or correctly detected based on their internal confidence score.

\subsection{Hallucination Reduction}

\begin{figure}[t!]
    \centering
    \includegraphics[width=\textwidth]{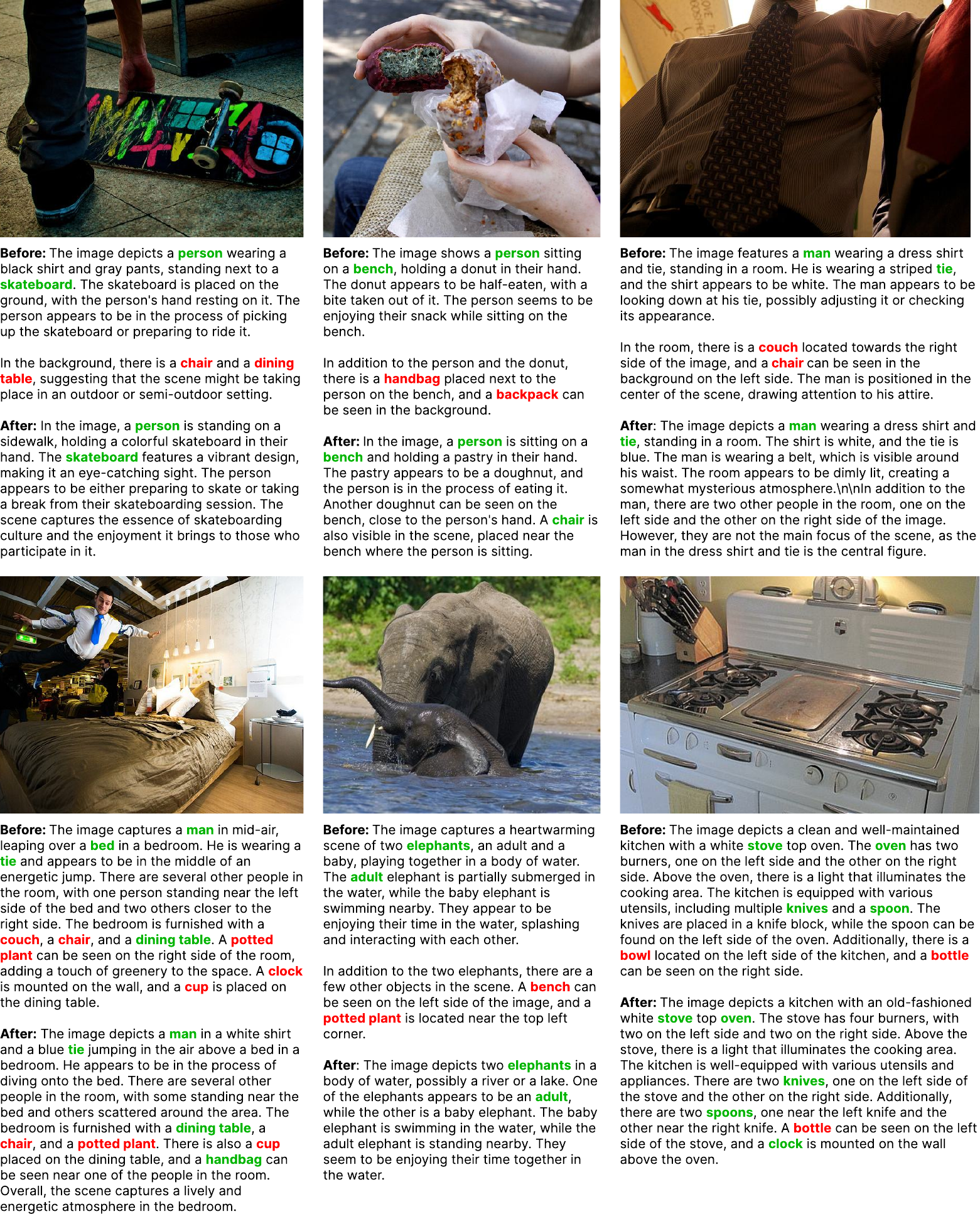}
    \vspace{-1em}
    \caption{\textbf{Qualitative results for LLaVA hallucination intervention}. Our algorithm removes hallucinations and, at times, adds correctly detected objects.}
    \label{figure:llava7b_hallucination_reduction_samples}
\end{figure}

\begin{figure}[t!]
    \centering
    \includegraphics[width=\textwidth]{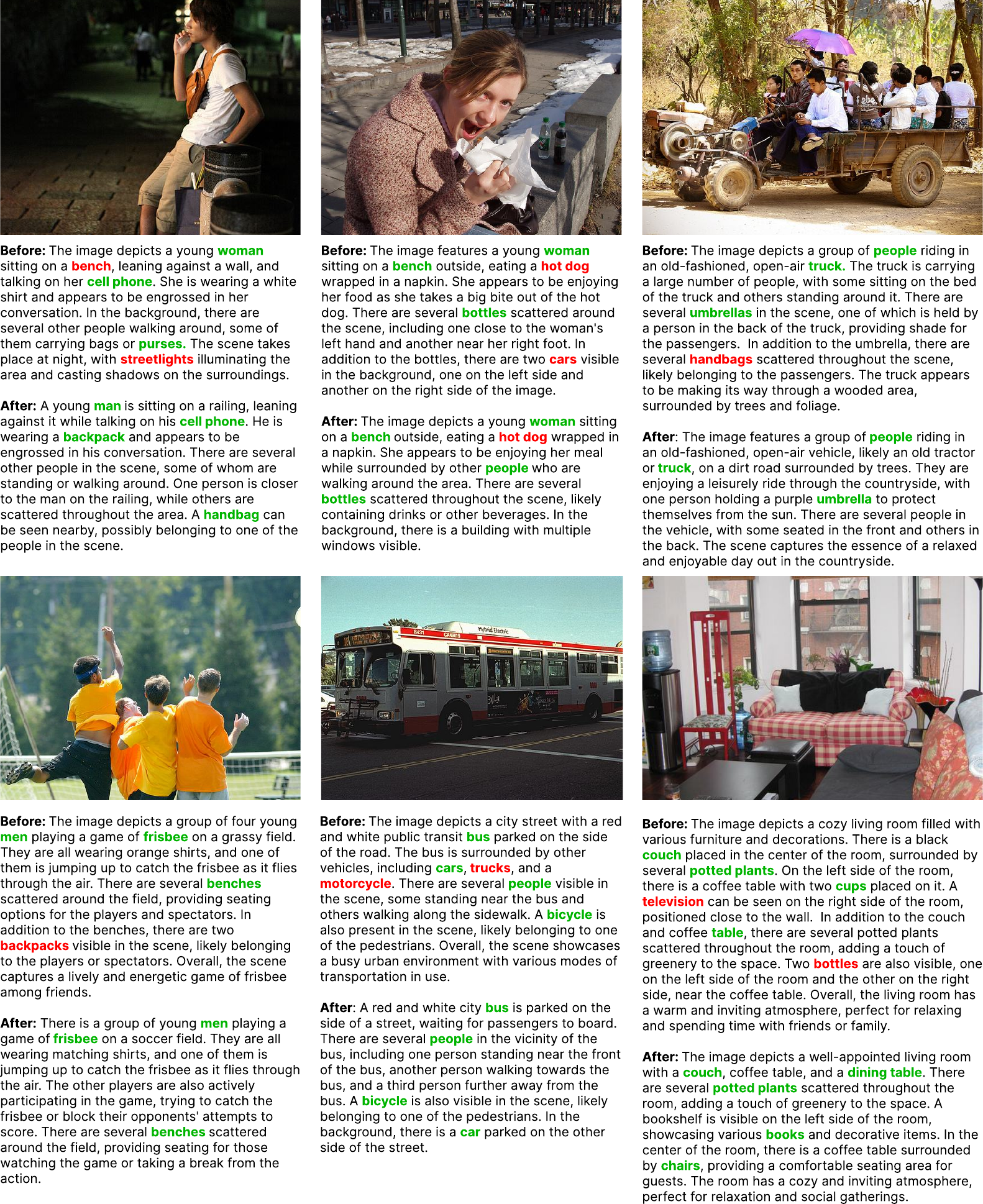}
    \vspace{-1em}
    \caption{\textbf{Qualitative results for InstructBLIP hallucination intervention}.}
    \label{figure:blip7b_hallucination_reduction_samples}
\end{figure}

We exhibit sample results from our hallucination reduction approach (\Cref{sec:hallucination_reduction}), which linearly removes text representations of hallucinations from image representations, in \Cref{figure:blip7b_hallucination_reduction_samples} for InstructBLIP and \Cref{figure:llava7b_hallucination_reduction_samples} for LLaVA. We show the image caption before and after our linear editing method, removing objects detected as hallucinations.

\subsection{Quantitative Evaluations on More Advanced Models}
\label{appendix:more_advanced_models}
We evaluate our approach on two additional models, LLaVA-NeXT 7B \citep{liu2024llavanext} and Cambrian-1 8B \citep{tong2024cambrian1fullyopenvisioncentric} with Llama 3. We threshold hallucinations as $c_o < 0.4$ for LLaVA-NeXT and $c_o < 0.3$ for Cambrian-1. Based on qualitative examples and referencing optimal parameters from other models in \Cref{Section 4.2}, we select $(l^I = 24, l^T = 22, \alpha = 2)$ for both models. We show quantitative results for hallucination detection in \Cref{table:more_hallucination_detection} and for hallucination intervention in \Cref{table:hallucination_removal_recent}. With our method, we observe a 27.73\% improvement in CHAIR$_S$ with LLaVA-NeXT and a 28.86\% improvement with Cambrian-1, demonstrating consistency with our findings on the LLaVA and InstructBLIP models.

\subsection{Object Localization}
\label{appendix:object_localization}
\begin{figure}[t!]
    \centering
    \includegraphics[width=\textwidth]{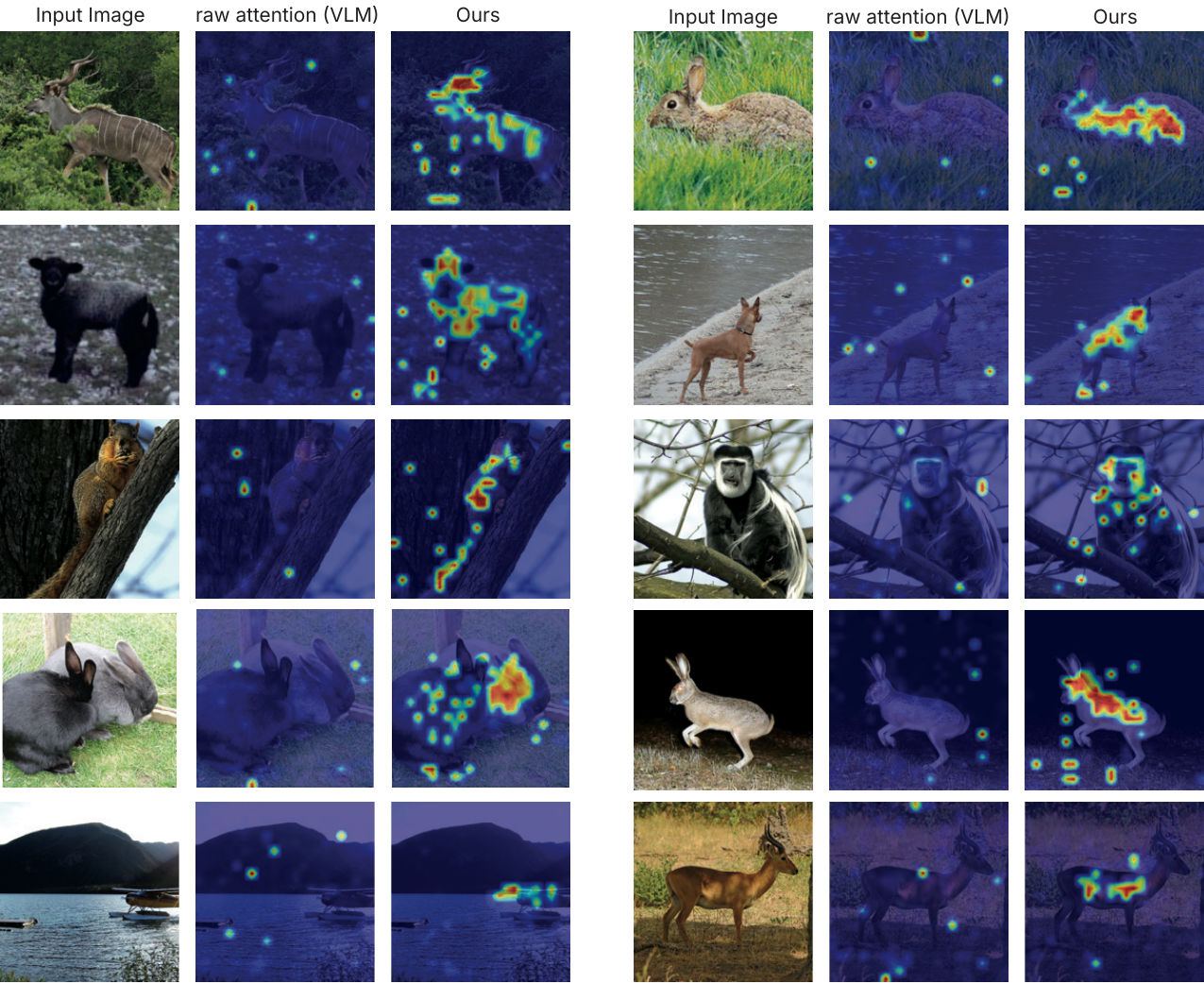}
    \vspace{-1em}
    \caption{\textbf{Object Localization Samples}.}
    \label{figure:qualitative_localization}
\end{figure}

We show qualitative examples for localization with internal confidence for specific image representations, specifically for the LLaVA model, in \Cref{figure:qualitative_localization}.

\subsection{Attribute Hallucinations}
\label{appendix:attribute_hallucinations}
\begin{figure}[t!]
    \centering
    \includegraphics[width=\textwidth]{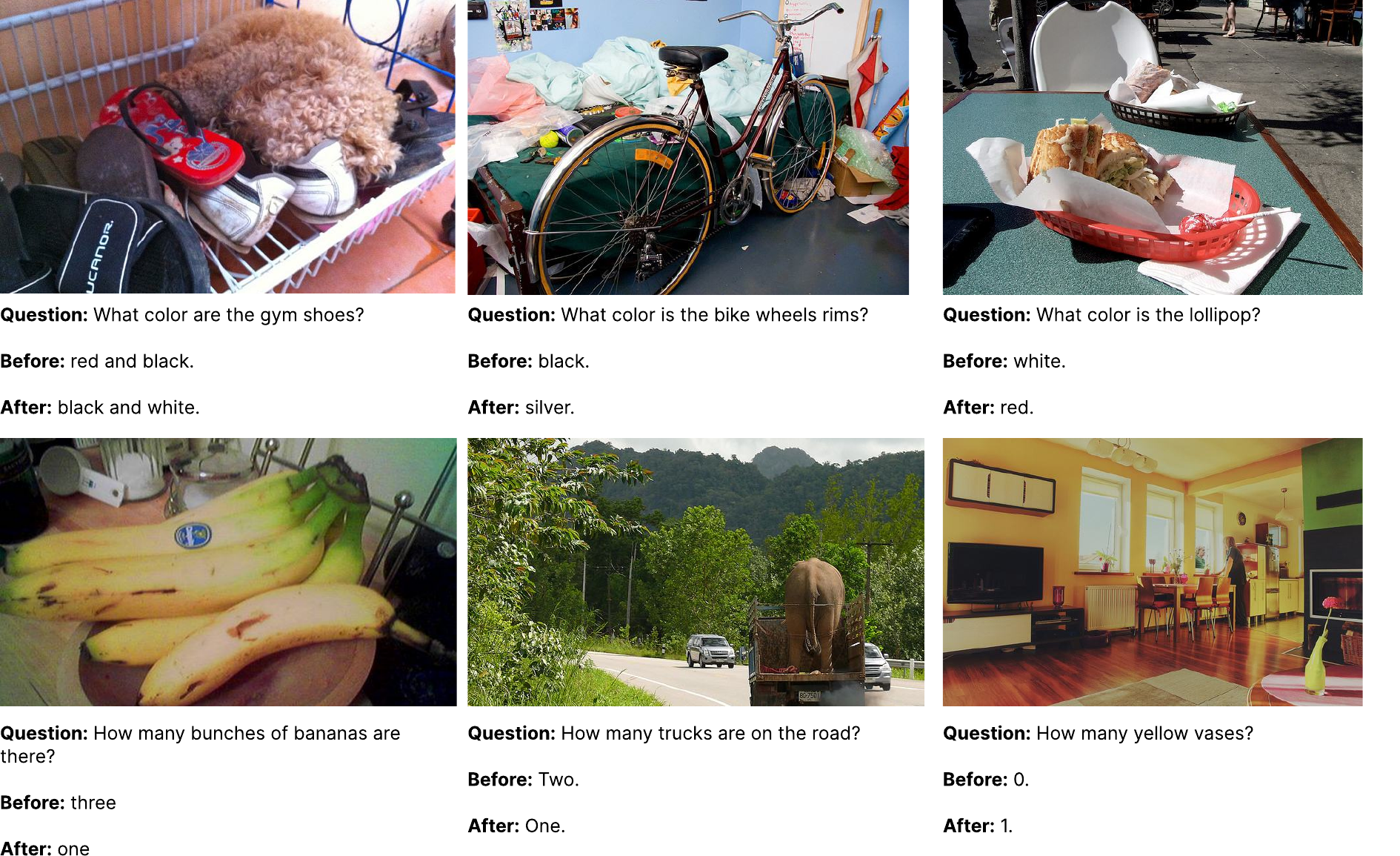}
    \vspace{-1em}
    \caption{\textbf{Qualitative results for attribute hallucinations using InstructBLIP}. We filter the VQA dataset for color and object number inaccuracies and correct answers with low confidence scores ($c_o < 0.05$) using \textsc{ProjectAway}. We reuse the same hyperparameters previously chosen for InstructBLIP ($l^I = 1, l^T = 2, \alpha = 1.5$).}
    \label{fig:attribute_hallucintions}
\end{figure}

Our analysis in this paper centered on object hallucinations because automated tooling and benchmarks for attribute (ex. shape, color, number) hallucinations are relatively sparse. However, we demonstrate the applicability of our editing technique on attribute hallucinations with qualitative examples filtered from the VQA 2.0 challenge in \Cref{fig:attribute_hallucintions}. We reuse the editing hyperparameters for InstructBLIP ($l_I = 1, l_T = 2, \alpha = 1.5$) and only edit attributes with $c_o < 0.05$.

\subsection{Zero-shot Classification}
\label{appendix:zero_shot_classification}

We evaluate the strength of internal confidence derived from the logit lens on image representations for classification of the COCO class within patches of the image. We use the COCO ground truth segmentations to find ground truth classes for image patches. We determine the accuracy of the rankings found from logit lens internal confidence scores to predict the class per patch and present our results in \Cref{table:topk_patch_class_accuracy}. We find that these values highly vary across classes, which we hypothesize is because certain classes such as ``person'' are represented with more specific tokens such as ``doctor'', ``skier'', ``girl'', etc. resulting in lower internal confidence for the tokens in ``person'' while other objects like ``toothbrush'', ``banana'', and ``broccoli'' are described in the same word as the COCO class.

\subsection{Qualitative examples beyond COCO 2014}
\label{sec:more_qualitative}

\begin{figure}[t!]
    \centering
    \includegraphics[width=\textwidth]{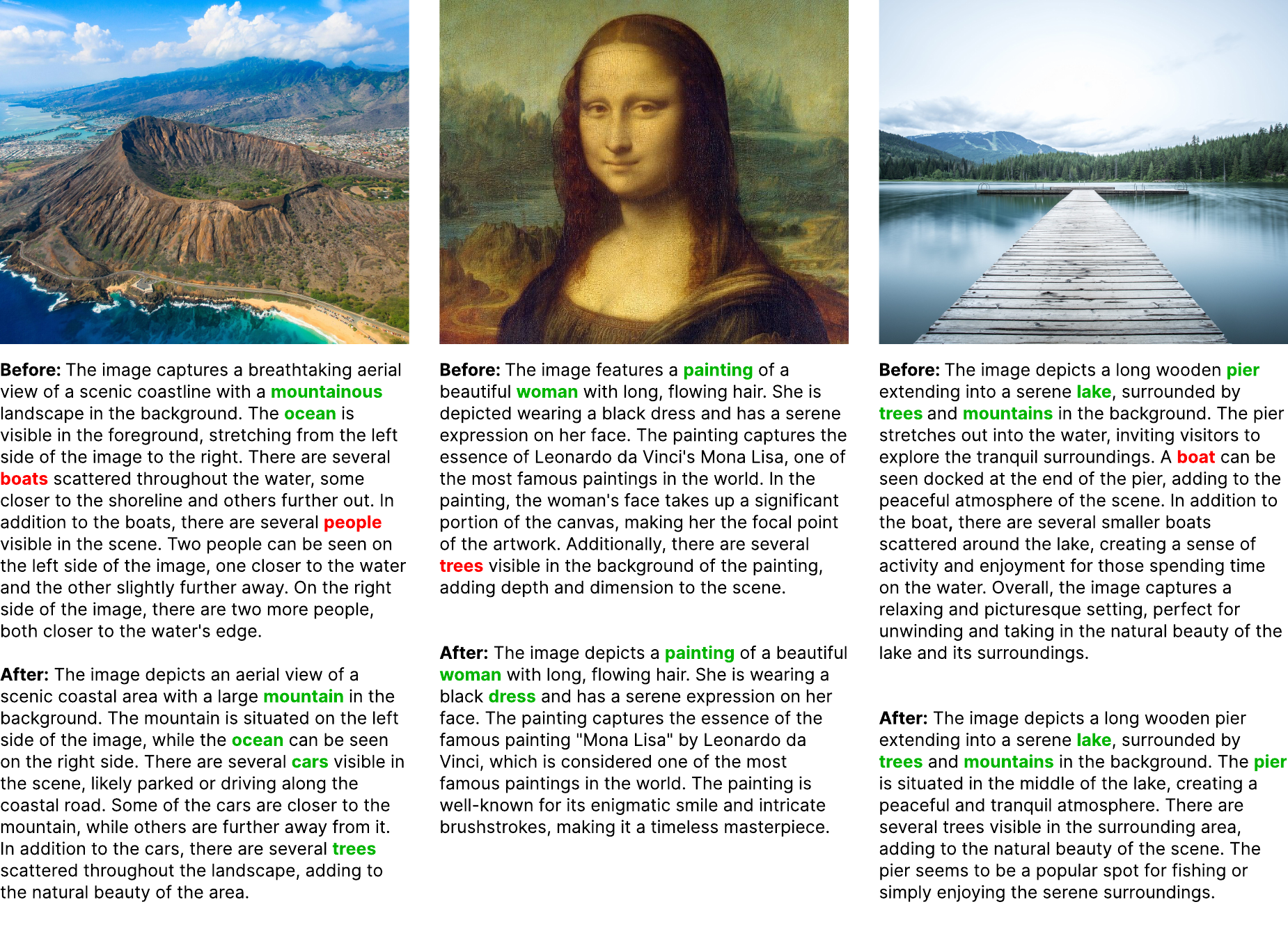}
    \vspace{-1em}
    \caption{\textbf{Qualitative results on images from LLaVA-Bench}. We randomly select images from the benchmark and use InstructBLIP to detect and edit out hallucinations. Our hyperparameter selection is the same as in \Cref{sec:sub:individual_removal} ($l^I = 1, l^T = 2, \alpha = 1.5$).}
    \label{fig:llava_bench_qual_examples}
\end{figure}

We focus on COCO 2014 in our analyses because CHAIR, our main evaluation criteria, is tied with the dataset and can automatically categorize objects of interest in image captions. While COCO 2014 is a diverse set of images, we provide qualitative examples of hallucination reduction (see \Cref{sec:hallucination_reduction}) on images from LLaVA-Bench~\citep{liu2024llavanext}, a collection of 24 images of varying subjects. The examples in \Cref{fig:llava_bench_qual_examples} using InstructBLIP align with the strong hallucination reduction observed with COCO 2014.

\begin{table}[t]
\caption{\textbf{Object presence classification performance.} We use internal confidence $c_o$ as a confidence score to classify whether the object is present in the image. We evaluate the mAP and ROC AUC of our classification method against the baseline for both the InstructBLIP and LLaVA models over a subset of 5000 COCO images.}
\begin{center}
\label{table:hallucination_detection}
\begin{tabular}{l|cc|cc}
\toprule
& \multicolumn{2}{c|}{\textbf{InstructBLIP}} & \multicolumn{2}{c}{\textbf{LLaVA}} \\
\textbf{Method} & \textbf{mAP} $\uparrow$ & \textbf{ROC AUC} $\uparrow$ & \textbf{mAP} $\uparrow$ & \textbf{ROC AUC} $\uparrow$ \\
\midrule 
Baseline & 0.53 & 0.55 & 0.49 & 0.47 \\
Ours & 0.78 & 0.83 & 0.60 & 0.68 \\
\bottomrule
\end{tabular}
\end{center}
\end{table}

\begin{table}[t]
\small
\caption{\textbf{Object presence classification on more models.} We classify whether the object is present in the image using internal confidence for LLaVA-NeXT and Cambrian-1 over a subset of 500 COCO images.}
\begin{center}
\label{table:more_hallucination_detection}
\begin{tabular}{l|cc|cc}
\toprule
& \multicolumn{2}{c|}{\textbf{LLaVA-NeXT}} & \multicolumn{2}{c}{\textbf{Cambrian-1}} \\
\textbf{Method} & \textbf{mAP} $\uparrow$ & \textbf{ROC AUC} $\uparrow$ & \textbf{mAP} $\uparrow$ & \textbf{ROC AUC} $\uparrow$ \\
\midrule 
Baseline & 0.93 & 0.66 & 0.94 & 0.73 \\
Ours & 0.95 & 0.75 & 0.97 & 0.83 \\
\bottomrule
\end{tabular}
\end{center}
\end{table}

\begin{table}[t]
\small
\caption{\textbf{Hallucination intervention performance on more models.} We mass-remove hallucinations detected by the method in \Cref{sec:hallucination_detection} on two more models, LLaVA-NeXT and Cambrian-1, on the same subset of 500 COCO images as used in \Cref{table:hallucination_removal}. We observe consistent improvement over the baseline while maintaining recall of objects present in the image.}
\begin{center}
\label{table:hallucination_removal_recent}
\begin{tabular}{lllll}
\toprule
\bf Model & \bf Method & \bf CHAIR$_i \downarrow$ & \bf CHAIR$_s \downarrow$ & \bf Recall (\%) $\uparrow$ \\
\midrule 
\multirow{2}{*}{LLaVA-NeXT}
& Beam Search & 6.8 & 23.8 & 63.12 \\
& Ours & \textbf{5.52} & \textbf{17.2} & 63.12 \\
\midrule
\multirow{2}{*}{Cambrian-1} & Beam Search & 3.27 & 9.2 & 53.28 \\
& Ours & \textbf{2.7} & \textbf{6.6} & 53.28 \\
\bottomrule
\end{tabular}
\end{center}
\end{table}

\begin{table}[t]
\small
\caption{\textbf{Per-class patch classification accuracy.} For each COCO class, we show the percentage of patches containing that object that top-k logit lens predictions can correctly identify.}
\begin{center}
\label{table:topk_patch_class_accuracy}
\begin{tabular}{@{}l@{\hspace{6pt}}r@{\hspace{6pt}}r@{\hspace{6pt}}r@{\hspace{6pt}}r@{\qquad}l@{\hspace{6pt}}r@{\hspace{6pt}}r@{\hspace{6pt}}r@{\hspace{6pt}}r@{}}
\toprule
\bf Class & \bf T3\% & \bf T5\% & \bf T10\% & \bf Patches & \bf Class & \bf T3\% & \bf T5\% & \bf T10\% & \bf Patches \\
\midrule
airplane & 51.0 & 61.8 & 72.5 & 102 & kite & 66.7 & 77.8 & 88.9 & 9 \\
apple & 70.2 & 76.6 & 80.9 & 47 & knife & 24.0 & 26.0 & 34.0 & 50 \\
backpack & 44.8 & 48.5 & 58.5 & 614 & laptop & 34.1 & 38.3 & 46.1 & 334 \\
banana & 77.2 & 79.2 & 83.2 & 101 & microwave & 23.8 & 37.2 & 55.2 & 223 \\
baseball bat & 0.0 & 33.3 & 33.3 & 3 & motorcycle & 64.7 & 70.6 & 78.3 & 984 \\
baseball glove & 65.4 & 69.2 & 73.1 & 26 & mouse & 50.0 & 50.0 & 66.7 & 6 \\
bear & 37.6 & 41.4 & 47.6 & 739 & orange & 44.2 & 50.0 & 57.7 & 52 \\
bed & 44.0 & 47.8 & 52.0 & 2373 & oven & 38.9 & 54.4 & 78.9 & 507 \\
bench & 61.3 & 63.9 & 66.6 & 524 & parking meter & 20.9 & 24.8 & 29.6 & 230 \\
bicycle & 27.2 & 32.3 & 58.3 & 235 & person & 0.5 & 1.1 & 12.6 & 11528 \\
bird & 75.5 & 79.4 & 80.6 & 155 & pizza & 41.8 & 52.1 & 69.4 & 3146 \\
boat & 26.2 & 29.7 & 38.8 & 516 & potted plant & 43.8 & 52.1 & 63.0 & 192 \\
book & 28.1 & 39.0 & 44.8 & 210 & refrigerator & 23.7 & 31.4 & 50.0 & 156 \\
bottle & 46.2 & 53.4 & 62.0 & 208 & remote & 14.3 & 17.3 & 17.3 & 98 \\
bowl & 22.1 & 25.0 & 31.9 & 1364 & sandwich & 40.2 & 44.4 & 54.3 & 468 \\
broccoli & 75.9 & 77.2 & 79.7 & 79 & scissors & 71.9 & 71.9 & 71.9 & 32 \\
bus & 49.8 & 54.1 & 61.3 & 1786 & sheep & 49.5 & 52.6 & 56.6 & 489 \\
cake & 43.2 & 51.4 & 83.5 & 1182 & sink & 57.0 & 60.3 & 65.1 & 272 \\
car & 29.4 & 37.7 & 52.7 & 714 & skateboard & 48.1 & 50.6 & 57.7 & 239 \\
carrot & 50.0 & 50.0 & 75.0 & 4 & skis & 27.6 & 34.2 & 44.7 & 76 \\
cat & 57.7 & 61.5 & 66.5 & 2239 & snowboard & 37.5 & 41.7 & 47.9 & 48 \\
cell phone & 73.4 & 76.6 & 82.8 & 64 & spoon & 35.5 & 43.5 & 53.2 & 62 \\
chair & 31.6 & 33.1 & 37.6 & 516 & sports ball & 4.4 & 8.9 & 20.0 & 45 \\
clock & 67.6 & 70.6 & 75.9 & 299 & stop sign & 85.7 & 88.6 & 89.9 & 237 \\
couch & 33.4 & 38.9 & 63.2 & 2435 & suitcase & 38.1 & 40.7 & 44.7 & 472 \\
cow & 51.9 & 58.6 & 67.0 & 324 & surfboard & 48.6 & 57.5 & 69.9 & 146 \\
cup & 9.4 & 14.9 & 30.4 & 181 & teddy bear & 38.5 & 43.1 & 47.7 & 239 \\
dining table & 25.4 & 47.6 & 74.9 & 7403 & tennis racket & 88.9 & 88.9 & 88.9 & 9 \\
dog & 45.9 & 51.3 & 59.7 & 1057 & tie & 46.2 & 49.7 & 52.8 & 197 \\
donut & 34.8 & 40.0 & 45.2 & 115 & toilet & 92.0 & 97.3 & 99.6 & 1131 \\
elephant & 47.0 & 57.4 & 64.9 & 902 & toothbrush & 78.9 & 78.9 & 100.0 & 19 \\
fire hydrant & 43.4 & 47.0 & 50.1 & 419 & traffic light & 45.5 & 45.5 & 45.5 & 11 \\
fork & 41.4 & 45.7 & 54.3 & 70 & train & 40.7 & 45.0 & 52.1 & 3008 \\
frisbee & 50.0 & 60.6 & 71.2 & 66 & truck & 34.5 & 40.1 & 54.6 & 930 \\
giraffe & 32.0 & 39.8 & 62.8 & 810 & tv & 27.2 & 32.6 & 37.6 & 298 \\
handbag & 42.2 & 53.0 & 57.8 & 83 & umbrella & 39.5 & 40.5 & 42.8 & 526 \\
horse & 63.7 & 65.8 & 68.8 & 240 & vase & 29.9 & 32.6 & 45.1 & 264 \\
hot dog & 56.9 & 62.7 & 69.3 & 394 & wine glass & 61.9 & 67.0 & 74.3 & 218 \\
keyboard & 43.1 & 47.7 & 49.2 & 65 & zebra & 33.0 & 35.9 & 44.5 & 373 \\
\midrule
\bf Overall & \bf 32.7 & \bf 39.3 & \bf 52.2 & \bf 55988 & & & & & \\
\bottomrule
\end{tabular}
\end{center}
\end{table}


\end{document}